\documentclass[sigconf]{acmart}

\usepackage{enumitem}
\usepackage{caption}
\usepackage{subcaption}
\usepackage{wrapfig}
\usepackage{multirow}
\usepackage{hhline}
\usepackage{tabularx,booktabs}

\AtBeginDocument{%
	\providecommand\BibTeX{{%
			\normalfont B\kern-0.5em{\scshape i\kern-0.25em b}\kern-0.8em\TeX}}}

\setcopyright{none} 
\acmDOI{}

\acmConference[]{Preprint}{version 2}{}
\acmPrice{}
\acmISBN{}

\begin{document}
	\pagestyle{plain}
	\title{Itsy Bitsy SpiderNet: Fully Connected Residual Network for Fraud Detection
	}
	
	\author{Sergey Afanasiev}
	\affiliation{%
		\institution{National Research University HSE}
		\city{Moscow}
		\country{Russia}}
	\email{ }
	
	\author{Anastasiya Smirnova}
	\affiliation{%
		\institution{National Research University HSE}
		\city{Moscow}
		\country{Russia}}
	\email{ }
	
	\author{Diana Kotereva}
	\affiliation{%
		\institution{National Research University HSE}
		\city{Moscow}
		\country{Russia}}
	\email{ }
	
	
	\begin{abstract}
		With the development of high technology, the scope of fraud is increasing, resulting in annual losses of billions of dollars worldwide. The preventive protection measures become obsolete and vulnerable over time, so effective detective tools are needed. In this paper, we propose a convolutional neural network architecture SpiderNet designed to solve fraud detection problems. We noticed that the principles of pooling and convolutional layers in neural networks are very similar to the way anti-fraud analysts work when conducting investigations. Moreover, the skip-connections used in neural networks make the usage of features of various power in anti-fraud models possible. Our experiments have shown that SpiderNet provides better quality compared to Random Forest and adapted for anti-fraud modeling problems 1D-CNN, 1D-DenseNet, F-DenseNet neural networks. We also propose new approaches for fraud feature engineering called B-tests and W-tests, which generalize the concepts of Benford's Law for fraud anomalies detection. Our results showed that B-tests and W-tests give a significant increase to the quality of our anti-fraud models. The SpiderNet code is available at \url{https://github.com/aasmirnova24/SpiderNet}
	\end{abstract}

	\ccsdesc[300]{Security and privacy~Usability in security and privacy}
	
	\keywords{neural networks, fraud detection, CNN, feature engineering}
	
	
	\maketitle
	
	\section{Introduction}
	The development of high technologies contributes not only
	to the growth of corporations and world economies but also
	to the development of fraud, which leads to losses of billions
	of dollars every year around the world.
	
	In 2018, eight Indian banks incurred \textdollar1.3 billion in
	losses in a fraud case involving Kingfisher Airlines founder
	Vijay Mallya\footnote{\url{https://www.theguardian.com/world/2020/apr/20/kingfisher-airlinestycoon-vijay-mallya-loses-appeal-extradition-india}}. In another case, the Agricultural Bank of
	China faced losses of 
	
	\begin{figure}[ht]
		\centering
		\vspace{4.5 ex} 
		\includegraphics[width=\linewidth]{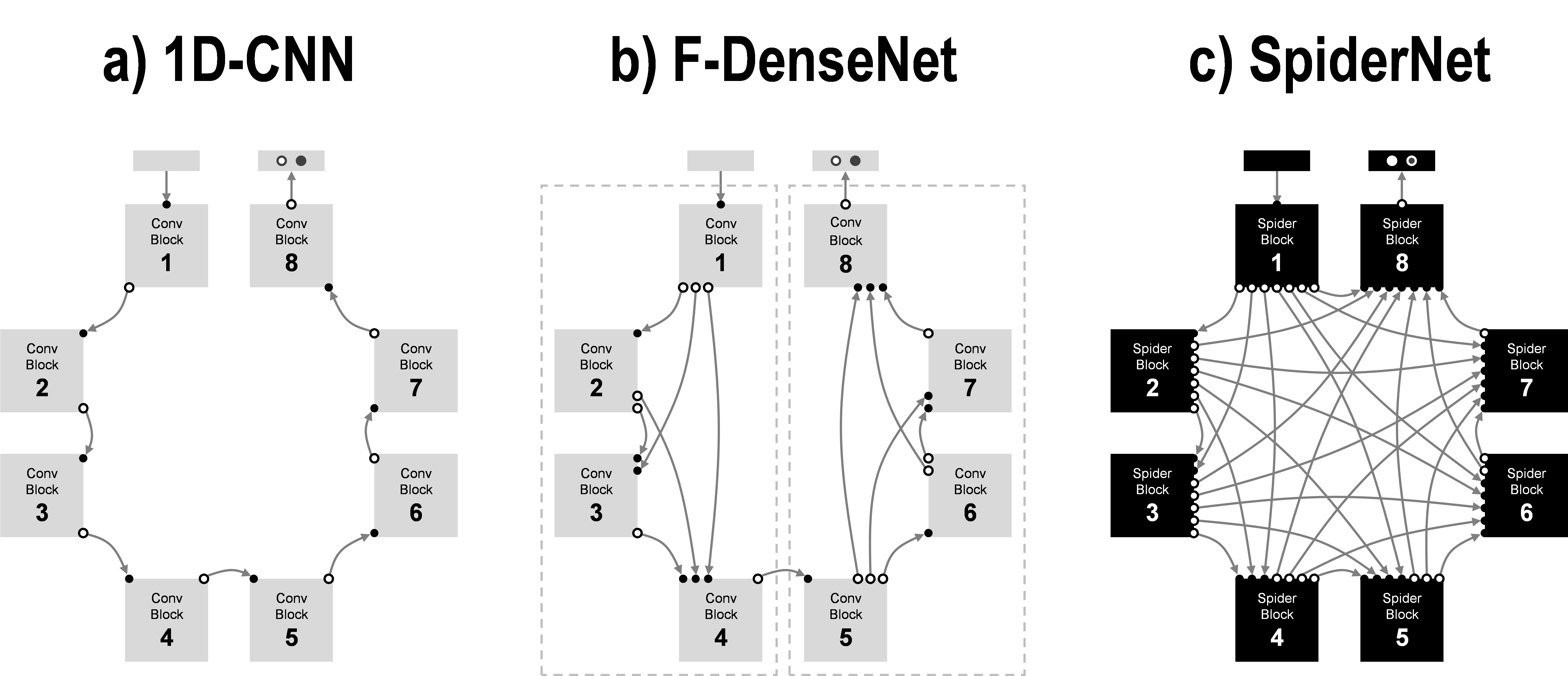}
		\caption{Convolutional neural network architectures designed for fraud detection: a) 1D-CNN, b) F-DenseNet, c) SpiderNet}
		\Description{Convolutional neural network architectures designed for fraud detection.}
	\end{figure} 
	
	\noindent\textdollar497 million after being defrauded by employees of billionaire Guo Wengui\footnote{\url{https://www.reuters.com/article/us-china-corruption-tycoonidUSKBN1900DL}}.
	
	Hacker attacks are another global problem. In 2019, the
	FBI issued an official announcement that global losses from
	fraudulent Business Email Compromise (BEC) reached \textdollar26
	billion during the period from June 2016 to July 2019\footnote{\url{https://www.ic3.gov/Media/Y2019/PSA190910}}.
	
	Another growing threat is social engineering, which has
	hit Russian bank customers seriously. According to the
	official data of the Bank of Russia, losses of Russian banks’
	clients from card fraud reached \textdollar130 million in 2020, which
	is 10 times higher than similar losses in 2017\footnote{\url{https://cbr.ru/analytics/ib/fincert/##a_119487}}.

	Anti-fraud tools can be roughly divided into directive,
	preventive, and detective. Directive tools such as instructions
	and warnings work like a scarecrow and only affect
	untrained fraudsters. Preventive tools help prevent fraud, but
	over time, fraudsters adapt and find ways to get around them.
	Detective tools are essential to detect fraud and minimize
	losses if fraud has not been prevented. Statistical approaches
	and machine learning methods are used to develop detective
	tools. However, there are unresolved problems
	in this area, such as instability and low generalizing ability
	of anti-fraud models, as well as high privacy of domain
	expertise\cite{Bolton06}.
	
	On the other hand, in recent years, we have seen outstanding advances in deep learning and the successful application of neural networks to practical tasks such as computer vision \cite{Krizhevsky28, He16, He17, Wang57} and natural language processing \cite{Graves15, TVan52, Rajpurkar44, Paulus42, Vaswani53}. This gives us hope that innovative ideas proposed in deep learning will help to remove some of the issues in fraud detection modeling.
	
	In this paper, we propose a convolutional neural network architecture SpiderNet designed to solve fraud detection problems. We noticed that convolutional and pooling layers principles are very similar to the methods of manual processing of information by anti-fraud analysts during investigations. In addition, skip-connections used in convolutional networks \cite{He16} make it possible to use features of various power, including fraud scores from external providers.
	
	Our proposed technique allows us to increase the generalizing ability of anti-fraud models and is an important advantage of SpiderNet over classical machine learning methods and popular neural network architectures. We show that SpiderNet provides better quality compared to Random Forest and convolutional 1D-CNN and F-DenseNet network architectures adapted for fraud modeling (Figure 1). Moreover, comparing the SpiderNet results with the classic CNN and 1D-DenseNet architectures, we show that the feature locality property is lost while working with tabular data but remains crucial for images. Therefore, the technique of transferring CNN architectures from the CV domain, which is popular in applied tasks, does not give the best result. This confirms the thesis that the application of neural networks to applied tasks requires a deep understanding of the domain area, and it is important to adapt neural network architecture to the specifics of the task.
	
	In addition to the SpiderNet architecture, in this paper, we propose new approaches for developing anti-fraud rules called B-tests and W-tests. Our feature engineering method is based on the idea of identifying manipulation in financial statements using Benford's law \cite{Benford04}. Our proposed approaches can be generalized to any data type, which allows developing B-tests and W-tests for any fraudulent schemes, and not just for identifying accounting manipulations, where Benford's law is applicable \cite{Lu37, Nigrini40, Lu38}. Our results showed that B-tests and W-tests give a significant increase to the quality of our fraud detection models.
	
	To assess the quality of fraud detection models, we use the AUC ROC and Average Precision (AUC PR) metrics, as well as developed by our team business metric PL (Prevented Losses) that allows estimating the funds saved from internal fraud. Our proposed PL metric allows us to solve an important industry issue of assessing the economic efficiency of models developed for industrial use.
	
	We train, evaluate, and compare our SpiderNet with other algorithms on two datasets – private and public. The private dataset contains data on loan applications and internal fraud by POS partners of a large Russian bank, one of the top 50 Russian banks in terms of assets. The public dataset was obtained from an online payment fraud detection competition organized by Ant Financial Services Group\footnote{\url{https://dc.cloud.alipay.com/index##/topic/data?id=4}} \footnote{\url{https://www.kaggle.com/gmhost/atec-anti-fraud/version/2}}.
	
	Testing SpiderNet on two datasets with different types of fraud (internal and transactional) gives us reason to believe that our proposed methods will work well for other types of fraud because SpiderNet is based on the general concepts of fraudulent behavior formulated by Edwin Sutherland and Donald Cressey in criminology \cite{Sutherland49} and Gary Becker in behavioral economics \cite{Becker03}.
	
	The rest of the paper is structured as follows:
	\begin{itemize}
		\item In section 2, we describe a general theoretical model of fraudulent behavior, provide a classification of anti-fraud tools, describe the principles of developing fraud detection models, and outline current problems in this area;
		\item In section 3, we provide a review of the current state of neural networks for image classification and fraud detection;
		\item In section 4, we describe the general intuition of the SpiderNet architecture and present the schema of the Spider-Block;
		\item In Section 5, we describe methods for the automated development of B-tests and W-tests anti-fraud rules;
		\item In Section 6, we describe our experiment: provide characteristics of datasets, feature engineering methods, data preprocessing methods, tricks for training models, hyperparameters  of algorithms, and metrics for models’ quality assessment;
		\item In section 7, we demonstrate the results of the experiments;
		\item And in section 8, we summarize, draw general conclusions and outline open questions for future research.
	\end{itemize}
	
	\section{Fraud Detection}
	The foundations of modern criminology and the theory of white-collar fraud were laid nearly 100 years ago by Edwin Sutherland and his student Donald Cressey \cite{Sutherland49},who are considered some of the most influential criminologists of the 20th century. Sutherland and Cressy proposed to consider the crime not from the position of criminal law, as was customary, but from the position of sociology, applying basic sociological concepts in criminology.
	
	Four decades later, Nobel laureate Gary Becker, using the principle of economics imperialism, described the economic model of crime \cite{Becker03}, according to which crime can be viewed as an activity that some people choose rationally, comparing the expected benefits and expected costs:
	\begin{equation}
		\left( 1 - \pi\right) \ast U\left( W_C\right)  - \pi \ast S > U\left( W_L\right)
	\end{equation}
	
	\begin{list}{}
		\item where  $\pi$ -- the probability of being caught (assessed by the criminal, i.e. subjectively);
		\item $U\left( \cdot \right)$ - individual utility function;
		\item $S$ -- penalties incurred in the event of capture (for example, a fine or criminal punishment);
		\item $W_C$ -– proceeds of crime;
		\item $W_L$ -– income from legal activities.
	\end{list}
	
	The left side of inequality (1) characterizes crime-related elements; the right side is the utility from legal earnings. Logically, when inequality is satisfied, the individual, other things being equal, will prefer to break the law.
	
	To minimize losses from fraudulent activities, companies need to develop and implement anti-fraud tools that affect the components of inequality (1). In particular, the total losses from fraud can be reduced by increasing the probability $\pi$ and decreasing the $W_C$ component by increasing fraudsters’ costs for bypassing anti-fraud protection.
	
	Anti-fraud actions can be divided into 3 types\footnote{The Information Security Standards (CISSP) use more granular typing: preventative, deterrent, detective, corrective, recovery, compensation, directive, administrative, logical/technical, physical.}:
	\begin{itemize}
		\item {\itshape Directive} –- instructions, regulations, training materials, contracts, etc.;
		\item {\itshape Preventive} –- tools that are aimed at preventing fraudulent activities (locks, safes, passwords, etc.);
		\item {\itshape Detective} –- tools used to detect fraud (anti-fraud rules and models, investigation techniques, etc.).
	\end{itemize}
	
	\begin{figure}[ht]
		\centering
		\includegraphics[width=\linewidth]{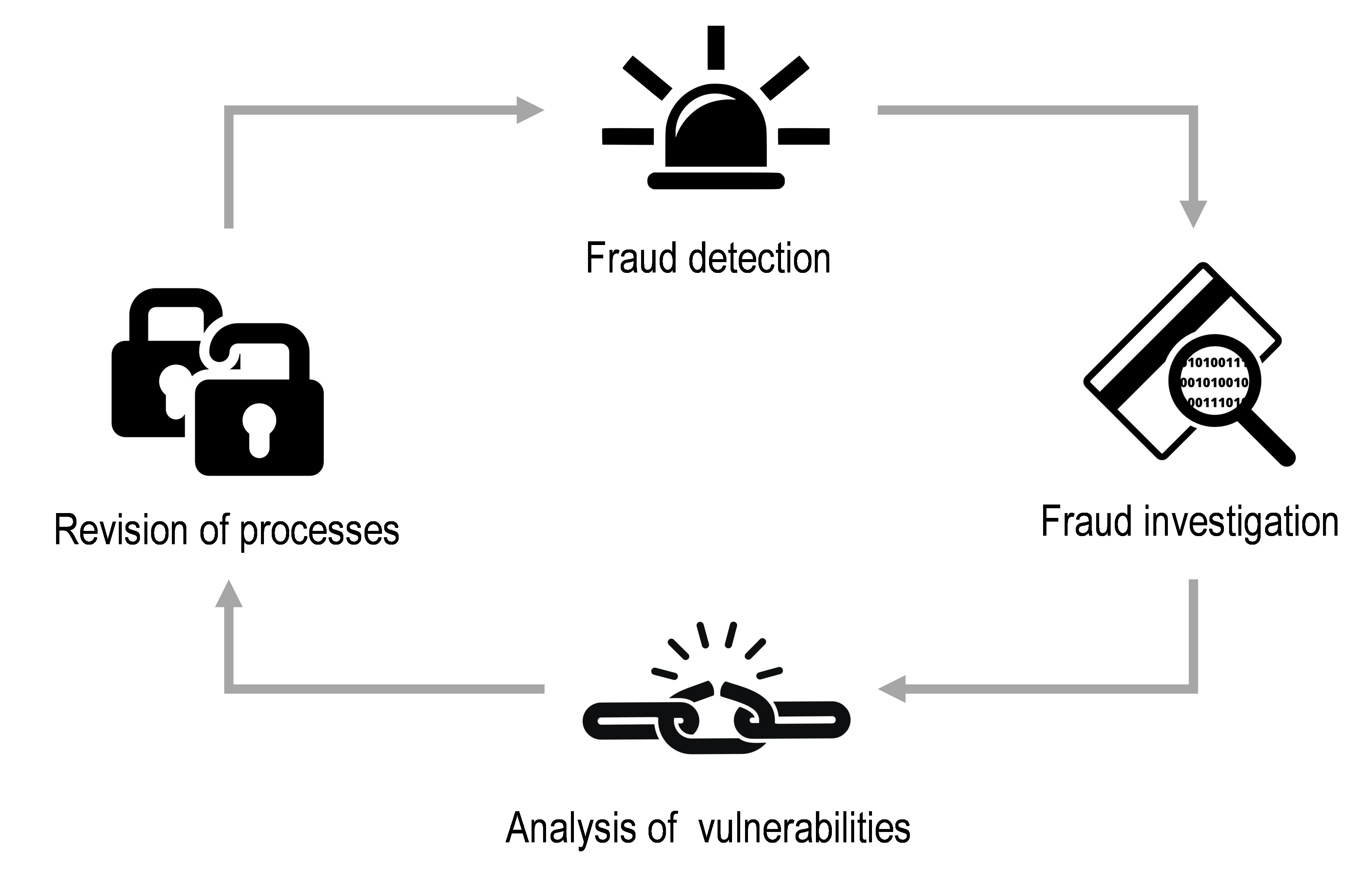}
		\caption{Arrangement of anti-fraud processes in the bank according to the principles of the scientific method.}
		\Description{Anti-fraud processes in the bank .}
	\end{figure}
	Preventive tools are the most effective, but fraud is highly adaptive, which means that fraudsters constantly come up with new ways how to get around the company's security. That is why the fight against fraud is compared to "confrontation of armor and projectiles". To face this constant arms race, the company's anti-fraud sector requires an integrated, systems approach. One of the main principles of the system approach is the arrangement of anti-fraud processes in the form of a cyclic scheme according to the principles of the scientific method.
	
	The scheme in Figure 2 shows how anti-fraud technologies are developed and improved. One of the main ideas of the scheme is that fraud and counteraction to fraud (anti-fraud) constantly influence each other \cite{Wallace54}. It means that, according to the scheme, the company should regularly review its anti-fraud processes, modifying and improving them. Therefore, detective tools that allow detecting new fraudulent schemes and vulnerabilities in processes are an important part of the anti-fraud system in a company.
	
	The detective tools are based on strong rules developed by experts and anti-fraud models built on these strong rules. Strong rules development consists of three key stages (Figure 3):
	\begin{enumerate}
		\item At the first stage, simple features and rules are developed by experts;
		\item At the second stage, complex rules are combined using arithmetic and logical operations over simple features and rules;
		\item At the third stage, strong rules with high predictive power for detecting fraud are selected from complex rules.
	\end{enumerate}
	The general principle of developing strong rules is similar to the method of collecting evidence in forensic science when all kinds of evidence are first collected, and then, based on the combinations of the collected evidence, a crime is proved. A similar principle is at the heart of convolutional neural networks, in which sequential convolution operations form complex features, and pooling operations filter out noise, i.e. select strong features that are good at predicting the target variable. This is why convolutional networks can be considered a powerful and intuitive algorithm for developing fraud detection models.

	\begin{figure}[ht]
		\vspace{4.5ex}
		\centering
		\includegraphics[width=\linewidth]{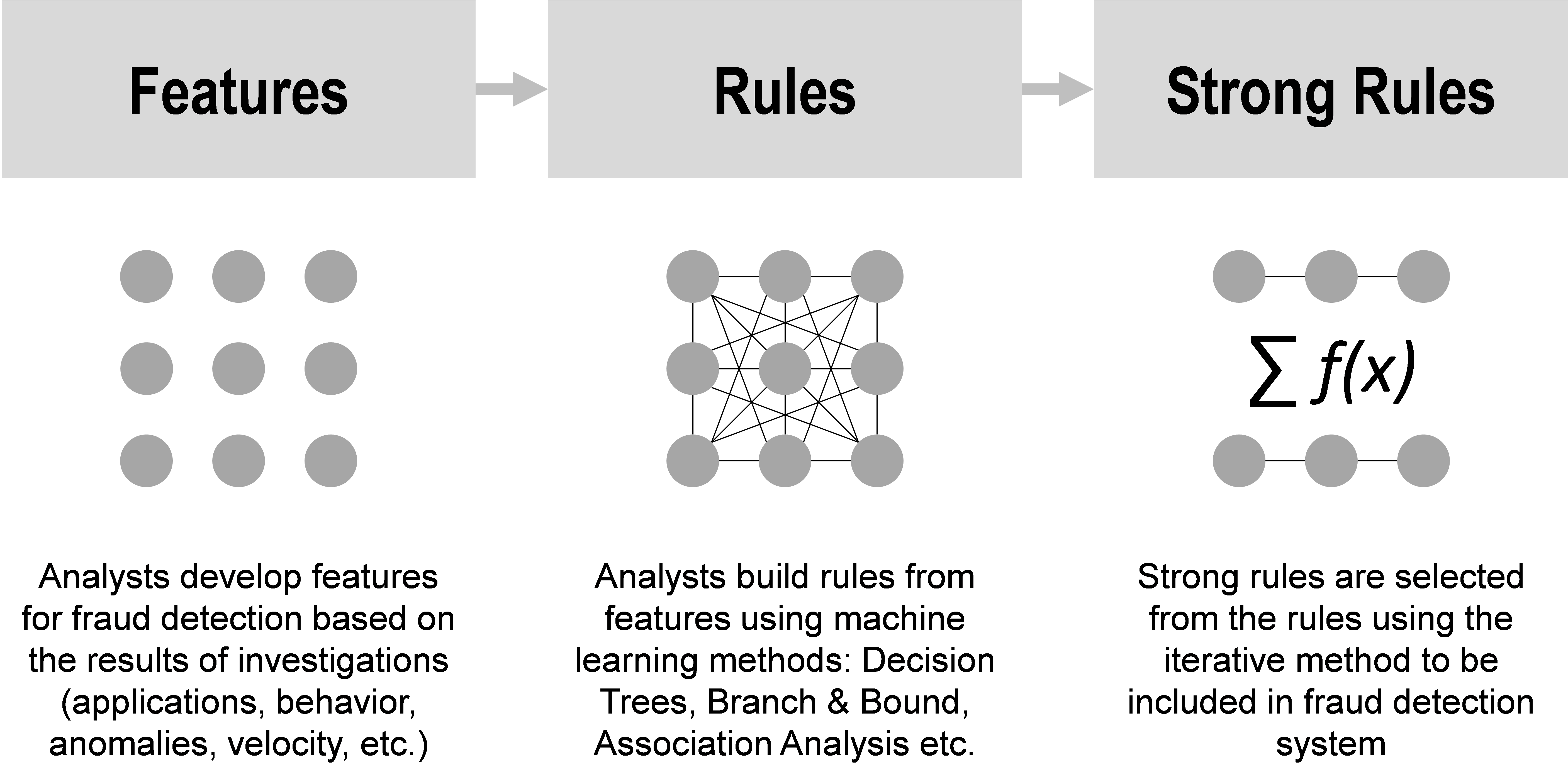}
		\vspace{1.3ex}
		\caption{Scheme of developing strong rules for fraud detection.}
		\Description{Scheme of developing strong rules for fraud detection.}
	\end{figure} 
	\section{Related work}
	\subsection{CNN Architectures}
	Over the past decade, convolutional neural networks (CNNs) have made a breakthrough in computer vision problems solution \cite{Paulus43}. The basic principles of CNN's work were borrowed from the works of Hubel and Wiesel, who studied the visual cortex in the 1950s and 1960s \cite{Hubel24, Hubel25}. The first CNN architecture was proposed by Yann LeCun in the late 1980s \cite{LeCun32}, and in the late 1990s, LeCun's research group developed the LeNet-5 architecture \cite{LeCun31}, which consisted of convolutional and pooling layers that perform the function of implicit regularization of the neural network.
	
	A breakthrough in computer vision came in 2012 when the deep convolutional neural network AlexNet won the ILSVRC computer vision competition in image classification problems \cite{Krizhevsky28}. The authors noticed that increasing CNN depth has an implicit regularization effect, which later became mainstream and was exploited in such architectures as VGG \cite{Simonyan48}, NiN \cite{Lin35}, Inception, GoogleNet \cite{Szegedy50}, ResNet \cite{He16}, and others.
	
	In 2014, the Google team showed excellent results at ILSVRC, proposing the GoogLeNet architecture \cite{Szegedy50}, a feature of which was Inception blocks with bottlenecks and 1x1 convolutions, which allowed increasing the number of convolutional channels, i.e. the width of the neural network. This technique also became popular and found its application in the architectures WRN \cite{Zagoruyko61}, Xception \cite{Chollet09}, ResNeXt \cite{Xie60}, MobileNets \cite{Howard21}, NASNet \cite{Zoph63, Zoph64}, etc.
	
	In 2015, the Microsoft Research team proposed another breakthrough architecture – ResNet \cite{He16} that used skip connections, which create an ensemble effect. This technique proved to be very effective and was later used in WRN \cite{Zagoruyko61}, Xception \cite{Chollet09}, ResNeXt \cite{Xie60}, FractalNet \cite{Larsson30}, DenseNet \cite{Huang22}, etc.
	
	In 2019, the Google Research team published an EfficientNet approach \cite{Tan51} to scale deep CNNs. Authors noticed that the key characteristics of convolutional architectures, such as depth, width, and resolution, depending on each other, therefore, for scaling, it is necessary to select the optimal combination of these parameters. Later, the EfficientNet approach allowed the participants to take the top places in the Deepfake Detection Challenge on Kaggle, which was organized by Facebook in early 2020.
	
	In addition to the developed architectures, effective tricks have been proposed for training CNN:
	\begin{itemize}
		\item Regularization method Weight Decay (L2 regularization) \cite{Krogh29, Jia26, He18};
		\item The Dropout technique \cite{Hinton20}, which allows to accidentally disconnect links in fully connected layers;
		\item Stochastic Depth technique \cite{Huang23} used in ResNet networks to random disable blocks;
		\item Augmentation methods: Cutout \cite{DeVries11}, Mixup \cite{Zhang62}, CutMix \cite{Yun46}, Valid/Diverse Noise, Flipping, etc. \cite{Xie59};
		\item Learning-rate reduction strategies that improve the convergence of gradient descent methods \cite{Donoghue41, Loshchilov36, Fort13, Li34}.
	\end{itemize}
	These and other proposed ideas allowed to achieve high results in Computer Vision.
	
	\subsection{Fraud Prediction Models}
	Statistical methods and machine learning have been used for fraud detection tools development for many years. In 2002, Bolton and Hand \cite{Bolton06} highlighted the key problems of using statistical methods in fraud detection modeling. Here are some of these issues that remain relevant today:
	
	\begin{enumerate}
		\item The scope of fraud is increasing with the development of high technology;
		\item Fraudsters bypass preventive technologies over time, so detective tools are needed;
		\item On the other hand, detective algorithms also degrade over time, so they need to be regularly updated;
		\item Databases and anti-fraud methods are closed from the scientific community. This makes it difficult to research and develop this area;
		\item To develop anti-fraud models, unsupervised (search for anomalies) and supervised (search for known fraudulent patterns) algorithms are used. Unsupervised models make a lot of mistakes, because anomalies may be caused by operational errors, marketing promotions, etc. Supervised models are trained on historical data, so they are poor at catching new fraudulent schemes;
		\item There is a global problem of class imbalance - there are much fewer fraudulent observations than non-fraudulent ones. This leads to a high false positive rate, which is why many false-positives are sent for investigation and the use of models in anti-fraud processes becomes expensive.
	\end{enumerate}
	In the discussion of this paper, Provost and Breiman noted other important issues:
	\begin{enumerate}[resume]
		\item Models are customized for a specific fraudulent scheme, which makes them difficult to scale to other types of fraud;
		\item Big data is needed to develop effective anti-fraud models, so this remains the lot of large companies;
		\item Machine learning algorithms do not solve the anti-fraud problem, expert knowledge and understanding of fraudulent schemes are needed.
	\end{enumerate}
	
	The development of anti-fraud models is most often solved as a binary classification problem (fraud/non-fraud), where classical algorithms are used, such as SVM, Logistic Regression, Random Forest, and others \cite{Bhattacharyya05}.
	
	With the onset of the deep learning boom, neural networks began to be used in fraud detection modeling and almost completely replaced classical machine learning methods in papers \cite{Kanika27}.
	
	Wiese and Omlin \cite{Wiese58} proposed using a recurrent LSTM network to detect fraudulent credit card transactions. The authors suggested that since recurrent networks were designed to process sequences, they should be good at detecting fraudulent patterns in card transaction sequences. Authors were able to demonstrate the advantage of the LSTM over the SVM algorithm, but the LSTM network failed to beat the simple fully connected neural network FFNN due to the insufficient number of fraudulent transactions in the dataset.
	
	Fu et al. \cite{Fu14} solved a similar problem in detecting card fraud by training the architecture of the LeNet-5 convolutional neural network developed by Yann LeCun for image processing. To train LeNet-5 on the card fraud detection problem, the authors presented transactions in the form of rectangular matrices of features, which were fed to the CNN input as two-dimensional pictures. The results obtained showed the superiority of LeNet-5 over the classical algorithms SVM, ANN, and Random Forest.
	
	Heryadi and Warnars \cite{Heryadi19} continued to develop these ideas and tried to combine CNN with a recurrent LSTM network. The authors' hypothesis was the following: CNNs, due to convolutions, should detect short-term fraudulent patterns, LSTM network, due to long short-term memory, should work well with long sequences of fraudulent transactions. Authors have developed a hybrid CNN-LSTM architecture, but experiments have shown that simple CNN detects fraud better than CNN-LSTM. Authors made an important conclusion from their results: long-term fraudulent schemes that cannot be detected for a long time are extremely rare, in contrast to short-term fast fraudulent schemes. This is confirmed by Becker's economic model of crime (eq. 1).
	
	Attempts to apply the popular neural network architectures CNN and RNN for fraud detection tasks continued in subsequent research efforts.
	
	Li et al. \cite{Li33} used the DenseNet architecture to detect electricity theft in China.
	
	Chen and Liu \cite{Chen07} refined the DenseNet architecture by adding an Inception module to the beginning of the network and additional skip connections between the Inception layer and Dense blocks. Their new CNN architectures named LI and DI have improved the results of standard CNN and DenseNet for the task of transaction fraud detection.
	
	Cheng et al. \cite{Cheng08} proposed a Spatio-temporal neural network STAN based on attention. The STAN architecture includes an Attention module and a simple CNN. For the task of detecting transaction fraud, STAN showed better quality compared to CNN, LSTM, etc.
	
	Li and Liu \cite{Zhenchuan65} proposed to use a special loss function for transaction fraud detection, which solved the problem of intraclass variability. Their loss function FCL (full center loss), constructed as a combination of DCL (distance center loss) and ACL (angle center loss), worked like batch normalization.
	
	For the tasks of organized fraud on Internet sites (fake reviews, bots, spam, etc.) detection, graph neural networks are gaining popularity today, allowing feature extraction for interconnected objects \cite{Shuhan47, Wang55, Dou12, Wang56}.
	
	\begin{figure}[ht]
		\vspace{5ex}
		\centering
		\includegraphics[width=0.985\linewidth]{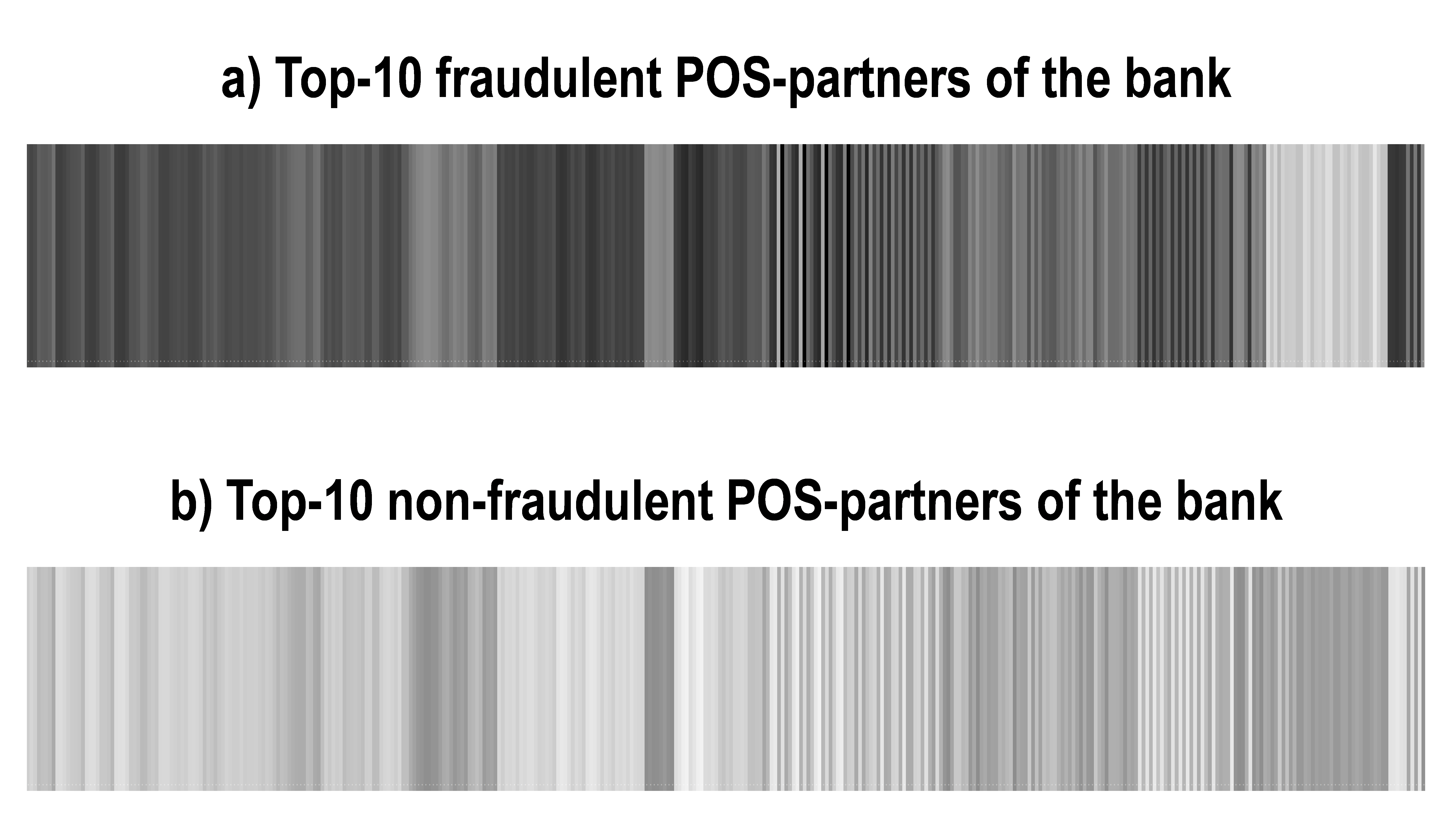}
		\vspace{1.3ex}
		\caption{Examples of one-dimensional heatmaps of features for internal fraud detection: a) average feature values for the top-10 fraudulent POS-partners of the bank; b) average feature values for the top-10 non-fraudulent POS-partners of the bank.}
		\Description{Examples of one-dimensional heatmaps of features.}
	\end{figure} 
	
	\section{SpiderNet}
	\subsection{Problem Formulation}
	One of the main problems of using neural networks in fraud detection tasks is that many of the proposed architectures were migrated from other domains (mainly from popular CV and NLP) without a deep understanding of why these architectures should work on fraud detection tasks.
	
	When designing a neural network architecture for fraud detection, our intuition is that anti-fraud rules developed by experts are a kind of digital evidence (by analogy with pixels in images, which are digital features processed by convolutions in CNN). At the same time, anti-fraud rules have different power like forensic evidence. Moreover, anti-fraud rules can work in conjunction with each other, strengthening the evidence base. This intuition tells us that CNN's convolution and pooling operations are the most appropriate tools for combining anti-fraud rules and selecting strong combinations of them.
	
	In computer vision problems, it was shown that a sequence of several convolutional layers allows one to create hierarchical feature maps using the locality property in images \cite{Lee66}. On the other hand, the locality property is lost in tabular data, since the features can be arranged in a different order. Nevertheless, studies of CNN architectures have shown that convolutions learn well not only medium-frequency (local) features but also low-frequency ones (texture, background, color, etc.) \cite{Goodfellow67, Hermann68}. This allows us to assume that CNN architectures should perform well on tabular data, where 1D convolutions can be applied to the feature vector. In classification problems, such feature vectors will differ for observations of different classes in color and texture (Figure 4).
	
	\begin{figure}[ht]
		\vspace{4.5ex}
		\centering
		\includegraphics[width=\linewidth]{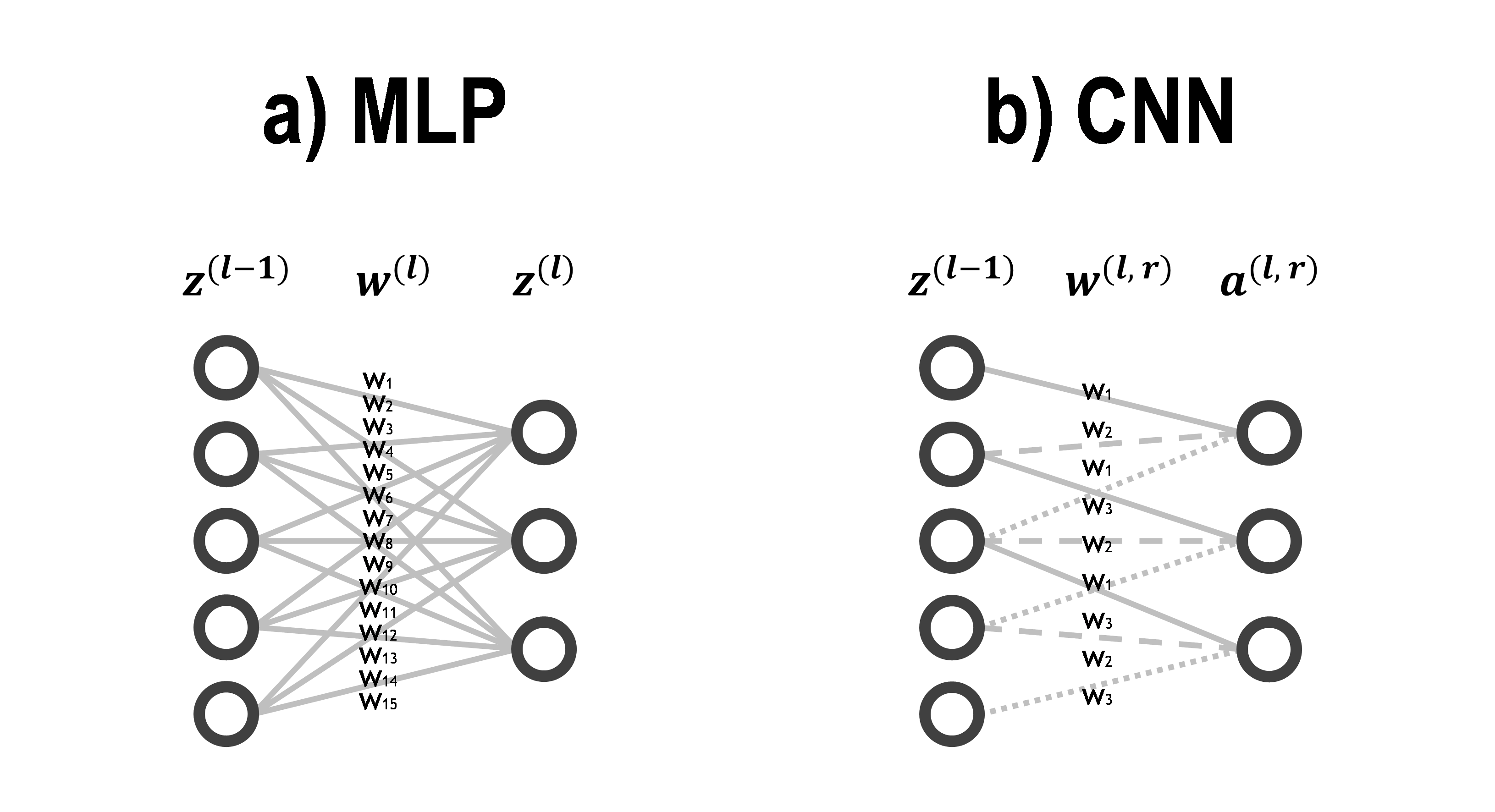}
		\vspace{1.3ex}
		\caption{The dropout principle between neurons in CNN: a) connections between layers in a fully connected neural network (MLP); b) connections between layers in a CNN.}
		\Description{The dropout principle between neurons in CNN.}
	\end{figure}
	
	The mathematical intuition of how convolutions work can be explained through CNN regularization by drop connections between neurons. Moreover, convolutions drop connections in an orderly manner, as opposed to the random dropout used in fully connected networks (Figure 5). In addition, the receptive field in convolutions allows capturing a small number of features, due to which many small models are trained in CNN on these subsets, which are then summed up, achieving the ensemble effect.
	
	On the other hand, if the combination of anti-fraud rules obtained on the hidden layers has a strong predictive ability, then we want to use this combination without additional processing, forwarding it to the output layer of the neural network using skip connections.
	
	This intuition well represents the best practices of anti-fraud investigations and can be implemented using a fully connected residual network, which we call SpiderNet (Figure 1).
	
	\subsection{Spider Block}
	SpiderNet consists of blocks that are connected using skip connections. Thus, each block receives features from all previous blocks, and these features are processed using convolutional and pooling layers and are forwarded to all subsequent blocks. Cause we want to have only the strongest features at the output, the blocks that are farthest from the network output contain several pooling layers, filtering out weak and medium features. Conversely, the closer the block is to the network output, the fewer pooling layers it contains since the features that have reached this block have already passed several convolutional and pooling layers. The general architecture of SpiderNet's block is shown in Figure 6.
	
	SpiderNet is a convolutional feedforward network with skip connections between blocks. Formally, the kth Spider-block is defined by the recursive formula:
	
	\begin{equation}
		y_k = \mathcal{F}_k\left( y_{k-1}\oplus\ldots\oplus y_{1}\right)  = \mathcal{F}_k\left(\sum_{i=1}^{k-1}\oplus y_i\right)
	\end{equation}
	
	\begin{list}{}
		\item where  $y_k$ is a vector of $n-k$ outputs for the kth block;
		\item $\mathcal{F}_k\left( \cdot \right)$ is the kth block operator, which combines the functions dropout, convolution, batch normalization, ReLU, and Max-pooling;
		\item $\oplus$ is the concatenation operator for incoming vectors;
	\end{list}
	\begin{figure}[ht]
		\centering
		\includegraphics[width=\linewidth]{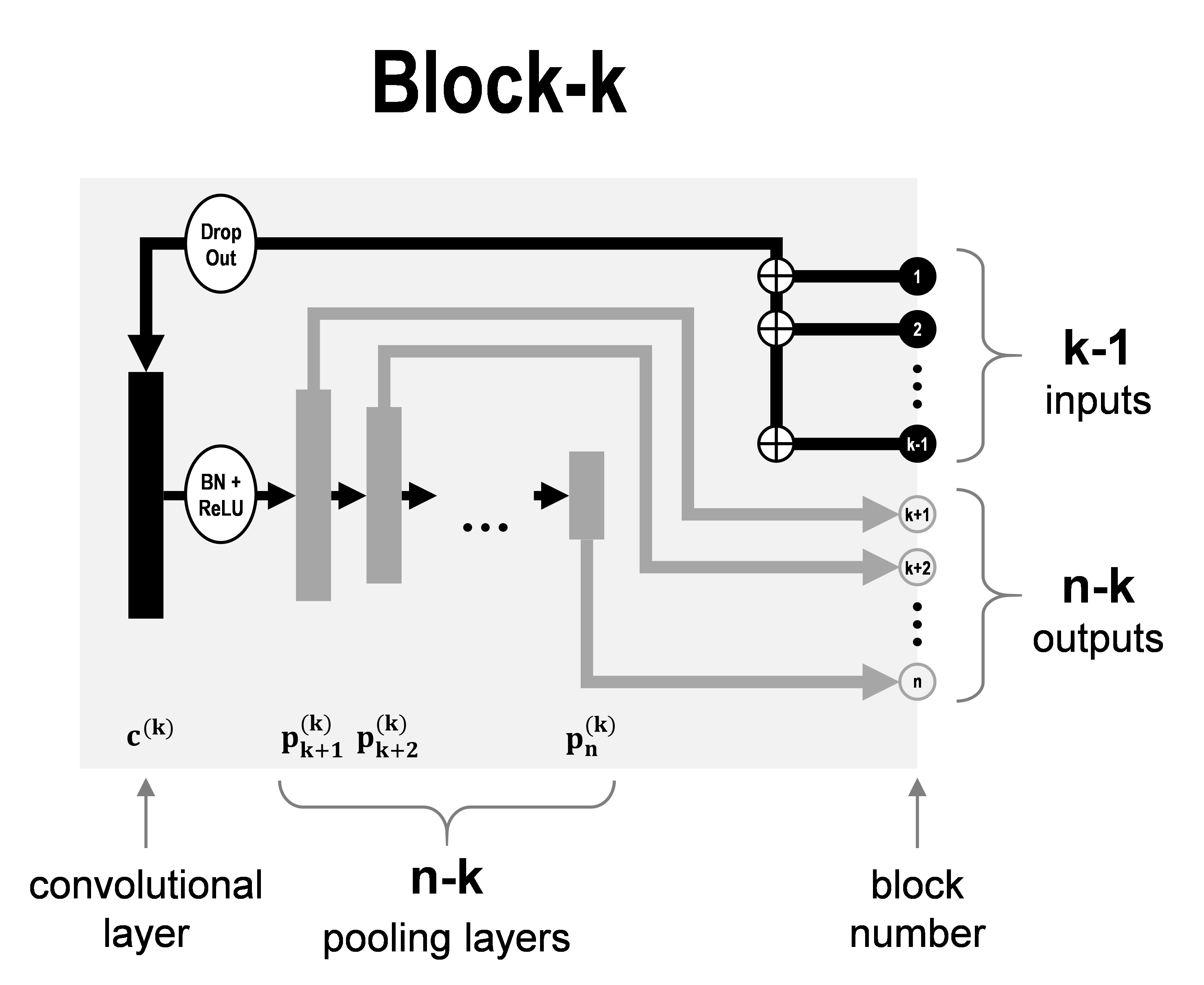}
		\caption{Scheme of the kth Spider-block with convolutional layer and $n-k$ pooling layers ($n$ is the total number of Spider-blocks).}
		\Description{Scheme of the kth Spider-block.}
	\end{figure}
	
	\begin{list}{}
		\item $\sum\oplus$ is a short notation for vector concatenation;
		\item $y_i$ is the output vector of the ith block, which is fed to the input of the kth block $\left(1 \leq i < k\right)$.
	\end{list}
	
	In the last nth Spider-block, after convolutional layer and BatchNorm+ReLU and MaxPooling layer, there is also global average pooling which is applied to reduce the dimension of the channels. After these operations, the vector $y_n$ is fed to two fully connected layers with dropout and SoftMax output for binary classification.
	
	Expression (2) demonstrates the mathematical intuition of the SpiderNet architecture. The output vector of the kth block is obtained by transforming the concatenated outputs from the previous blocks, which are smaller neural networks, which receive concatenated vectors of the previous blocks as input, etc. Thus, SpiderNet works as an ensemble of neural networks, which allows improving the convergence and generalization of the entire neural network.
	
	\section{B-tests and W-tests}
	The peculiarity of fraud modeling is that most of the fraud rules are developed manually by experts. This because fraudulent schemes are diverse, especially internal fraud schemes that are designed for specific vulnerabilities of the organization.
	
	We offer B-tests and W-tests approaches that automate the manual process and generalize feature engineering for various types of internal fraud.
	
	\subsection{B-tests}
	B-tests are based on Benford's law, discovered at the end of the 19th century by Simon Newcomb \cite{Newcomb39}, who noticed that the values of some numerical data start with one more often than two, two more often than three, and so on. Later, Frank Benford \cite{Benford04} empirically confirmed this law for various social and physical phenomena, and the first-digit law was named after him.
	
	In the 90s of the twentieth century, Mark Nigrini used Benford's law to audit financial statements and managed to identify managerial embezzlement of \textdollar2 million in the office of the Arizona State Treasurer \cite{Nigrini40}.
	
	Summarizing these ideas for all types of data (numerical and categorical), we propose a B-tests technique, which consists of comparing the distributions of data characterizing an object with the general population of all objects (for example, the activities of employees or partners of the company).
	
	The general idea is that internal fraud is rare, i.e. it does not greatly affect the distribution of the general population, while the data on the activities of fraudsters are very different from the average (Figure 7).
	
	B-tests can be customized according to various characteristics of the sample, such as the partner sphere of business, the region of the partner's location, the number of quantiles in distributions, the divergence cut-off threshold, etc. Thus, the setting of B-tests can be reduced to an optimization task of searching for anomalies \cite{Btest01}. 
	
	B-tests can be calculated using various metrics reflecting the divergence between two distributions, for example, Chi-squared test, K–S test, Anderson–Darling test, etc.
	
	\begin{figure}
		\centering
		\begin{subfigure}[b]{0.5\linewidth}
			\centering\includegraphics[width=\linewidth]{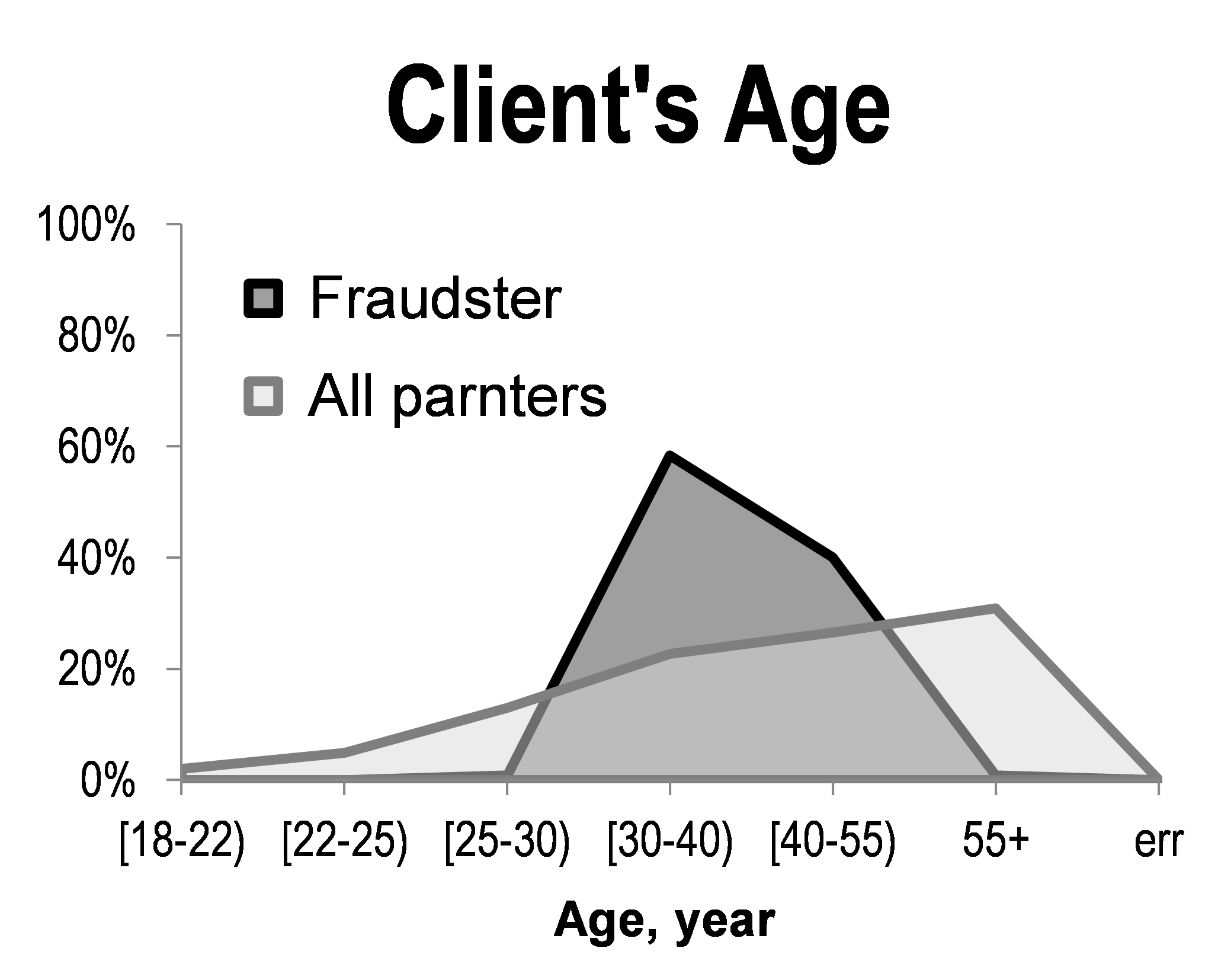}
		\end{subfigure}%
		\begin{subfigure}[b]{0.5\linewidth}
			\centering\includegraphics[width=\linewidth]{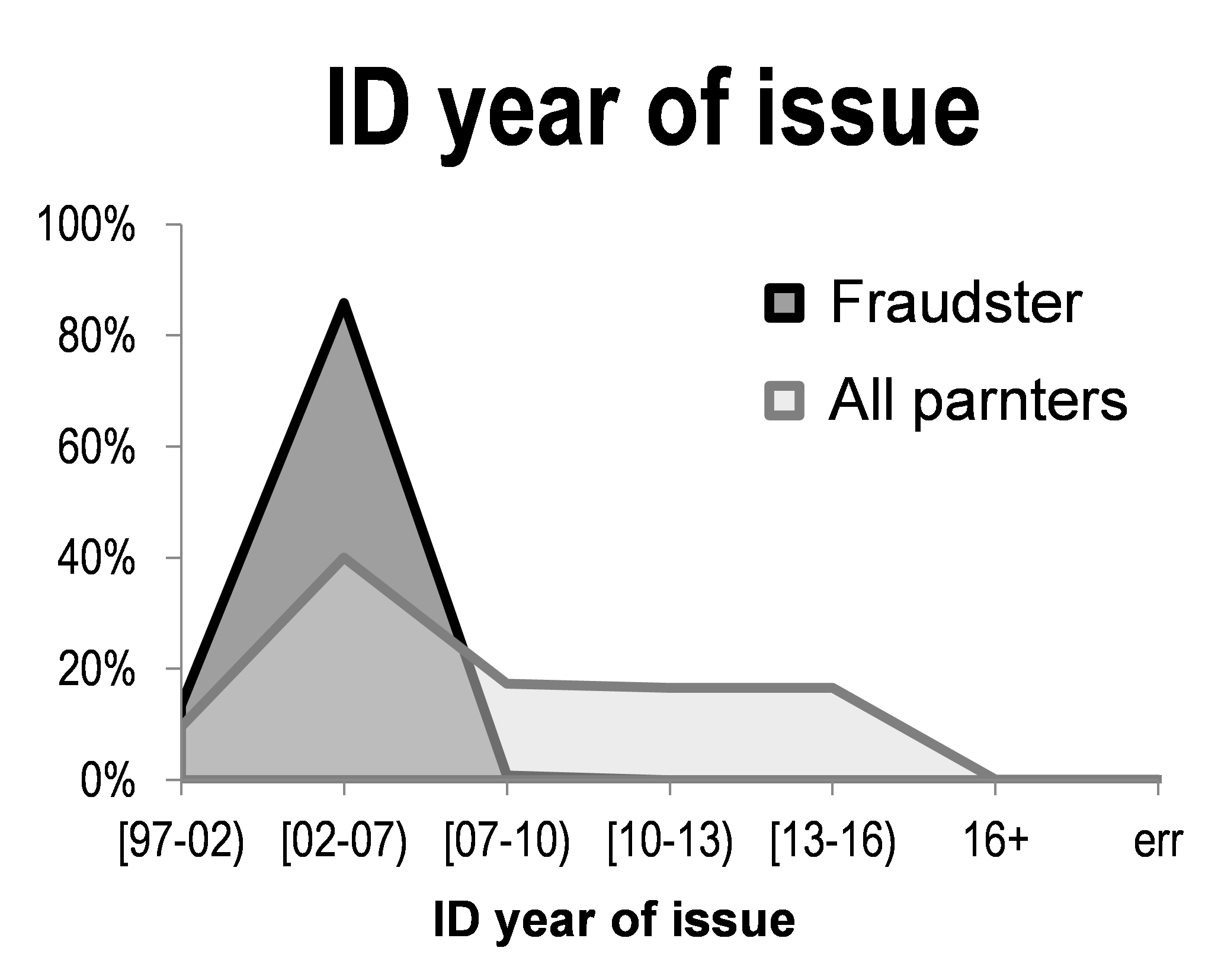}
		\end{subfigure}\vspace{0pt}
		
		\begin{subfigure}[b]{0.5\linewidth}
			\centering\includegraphics[width=\linewidth]{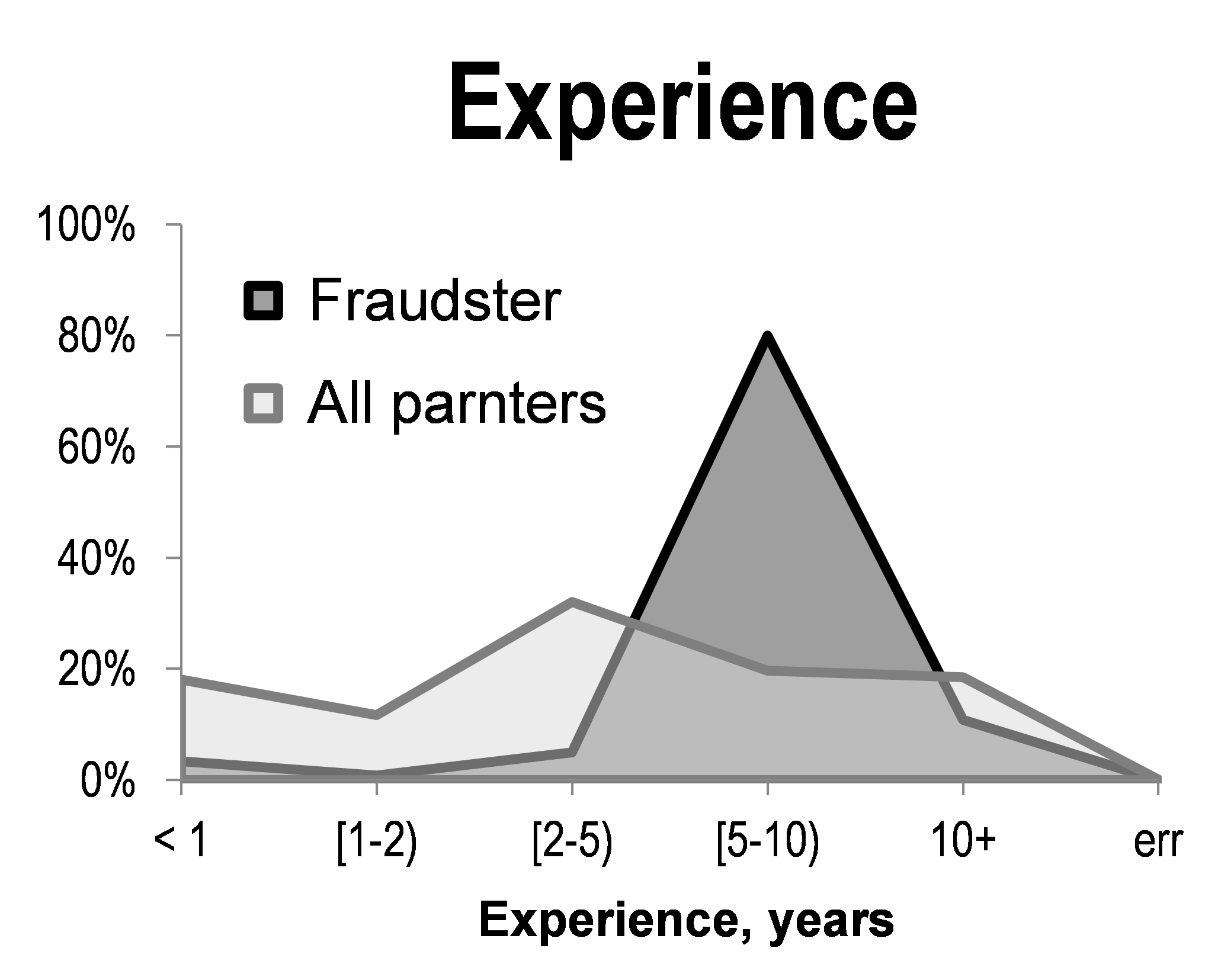}
		\end{subfigure}%
		\begin{subfigure}[b]{0.5\linewidth}
			\centering\includegraphics[width=\linewidth]{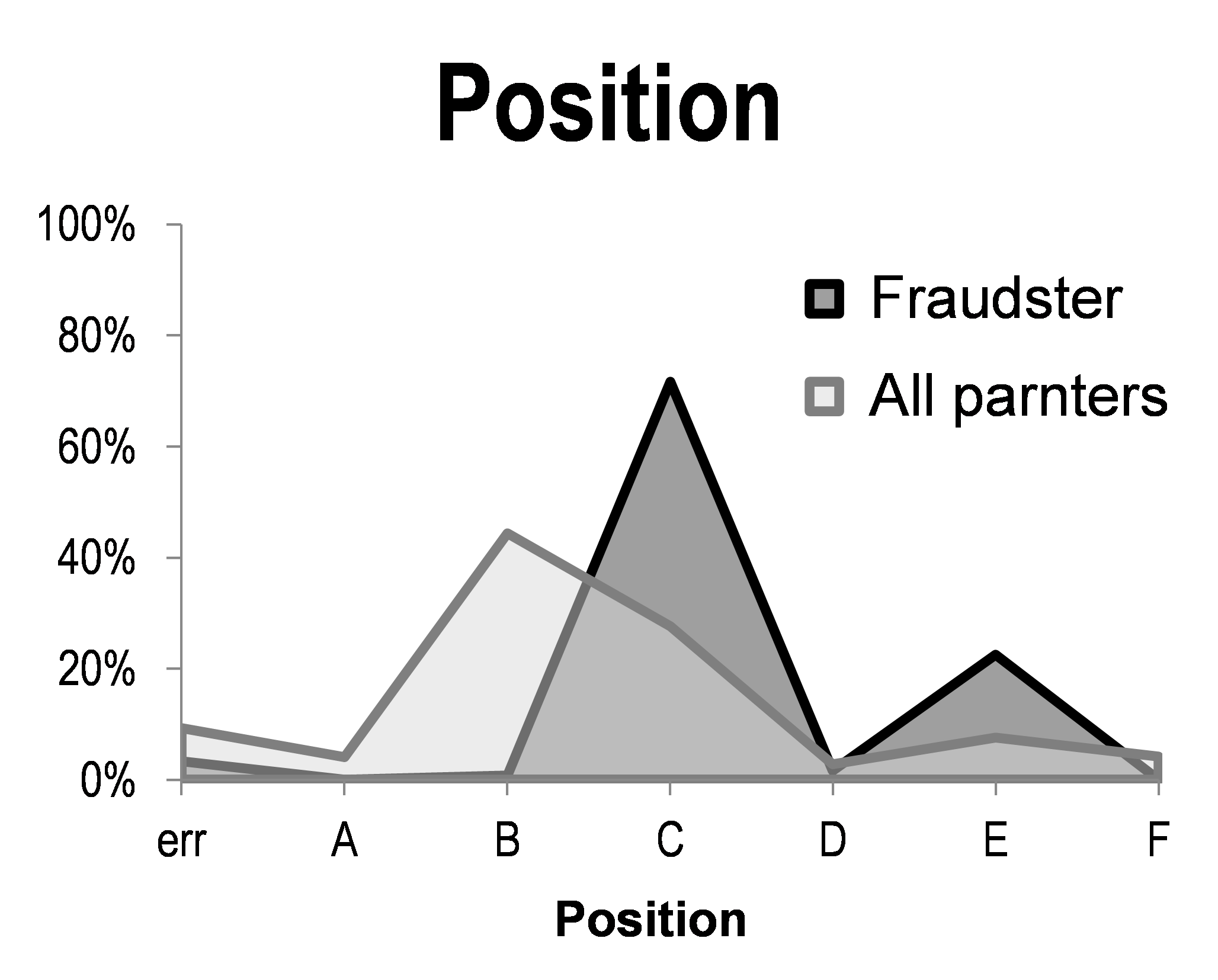}
		\end{subfigure}
		\caption{Examples of B-tests for internal fraud detection.}
		\Description{Examples of B-tests.}
	\end{figure}
	For our B tests, we calculate the area difference for two discrete distributions using the following formula:
	
	\begin{equation}
		S = \dfrac{1}{2} \sum_{i=1}^{n}\left| a_i - b_i\right| 
	\end{equation}
	
	\begin{list}{}
		\item where $a_i$ and $b_i$  are the compared distributions;
		\item $n$ is the number of quantiles in the distribution.
	\end{list}

	The number of quantiles in the distribution n and the threshold for S depend on the number of objects in the samples and are tunable hyperparameters.
	
	\subsection{W-tests}
	W-tests solve the same problem as B-tests using the Wasserstein metric, which is defined as:
	
	\begin{equation}
		W_p\left(\mu, \nu\right) = \inf_{\gamma \in \Gamma\left(\mu, \nu\right)} E_{\left(x, y\right) \sim \gamma} \left[ \left\| x - y\right\| \right] 
	\end{equation}
	
	\begin{list}{}
		\item where $E\left[Z\right] $ denotes the expected value of a random variable $Z$ and the infimum is taken over all joint distributions of the random variables $X$ and $Y$ with marginals $\mu$ and $\nu$  respectively.
	\end{list}
	Wasserstein metric solves the problem of the insufficient number of objects in samples, which is typical for B-tests. However, the Wasserstein metric can be calculated only for numeric data types, so W-tests cannot completely replace B-tests.
	
	\section{Experiment Setup}
	\subsection{Datasets}
	We trained and compared SpiderNet with other algorithms on two datasets: private data and public data. The general characteristics of datasets are presented in Table 1.
	
	{\itshape Private dataset}. Our private dataset contains data on the credit activity of POS partners of a large Russian bank, one of the top 50 Russian banks in terms of assets.
	Before applying the models, the bank used a rule-based approach. When the rules were triggered, the top-worst POS partners were sent to the bank's security for investigation. Since the security staff is limited, the number of investigations is also limited and amounts to 40 investigations per week. Thus, the main goal of applying the model approach was to reduce the time it takes to identify fraudulent partners with an unchanged investigation budget.
	
	The bank has a procedure for blocking unprofitable POS partners based on risk indicators, which are calculated approximately 90 days after the first fraudulent loan is issued. Therefore, to form the target variable, the loss for the POS partner was calculated within 90 days from the date of the fraud rules calculation. Unprofitable POS partners were marked as fraudulent, profitable ones - as non-fraudulent. This approach to constructing the target variable makes it possible to detect problem partners.
	
	Since fraud rules could have been triggered during the loss assessment period, the evaluation of the model approach will show the uplift to the existing rule-based process, and not the full economic effect of the model.
	
	The records in the private dataset are presented as a vector of features for each POS partner as of the weekly slice date. Therefore, a single POS partner can be included in the sample several times with different slice dates. The depth of calculation of the features ranges from 7 to 60 days ago from the date of the slice. A flag, indicating an internal fraud event, was set to each record as a target variable. The overall task comes down to predicting the internal fraud of the POS partner based on the data of its historical activities.
	
	{\itshape Public dataset}. The public dataset comes from an online payment fraud detection competition hosted by Ant Financial Services Group. The dataset contains the values of features for payment transactions and a binary target variable that reflects whether the payment is fraudulent or not.
	
	\begin{table}
		\begin{wrapfigure}{r}{0.05\linewidth} 
			\vspace{6.5ex}
			\hspace{-40ex}
			\includegraphics[width=\linewidth]{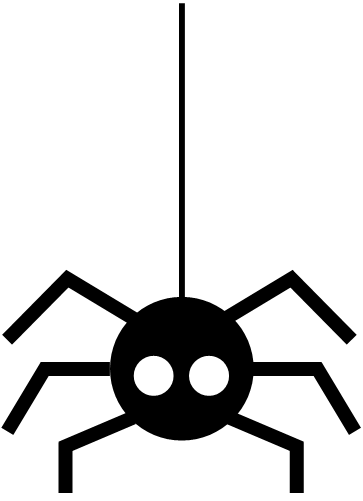}
		\end{wrapfigure}
		\begin{flushleft}
			\caption{Characteristics of private and public datasets}
			\label{tab:freq}
			\begin{tabular}{lll}
				\toprule
				Attribute&Private Data&Public Data\\
				\midrule
				{\bfseries Source} & Russian Bank& \parbox[l]{2cm}{Ant Financial Services Group}\\
				{\bfseries Type of data}& POS credits& Payments\\
				{\bfseries Type of fraud}&  Internal& Transaction\\
				{\bfseries Period time}& 03.2014-10.2019& 09.2017-11.2017\\
				{\bfseries All observations, \#}&  	1 880 499& 990 006\\
				{\bfseries Fraud observations, \#}&  5 327& 12 122\\
				{\bfseries Fraud rate, \%}&  0.28& 1.22\\
				{\bfseries All features, \#}&  509& 297\\
				{\bfseries Selected features, \#}&  163& 128\\
				\bottomrule
			\end{tabular}
		\end{flushleft}
	\end{table}

	\subsection{Feature Selection}
	To form the final samples, we did data preprocessing. Firstly, we checked the features for validity and removed features with a low fill rate. Secondly, we selected features using a cross-correlation matrix. Of a pair of highly correlated features, we left the one that had a stronger correlation with the target variable.
	
	As a result of data preprocessing, 163 features remained in the private dataset, and 128 features remained in the public dataset.
	
	\subsection{Model Architectures and Tricks}
	To assess the generalization ability and demonstrate the benefits of SpiderNet, we compared the results of several machine learning algorithms:
	\begin{figure*}[ht]
		\centering
		\includegraphics[width=\textwidth]{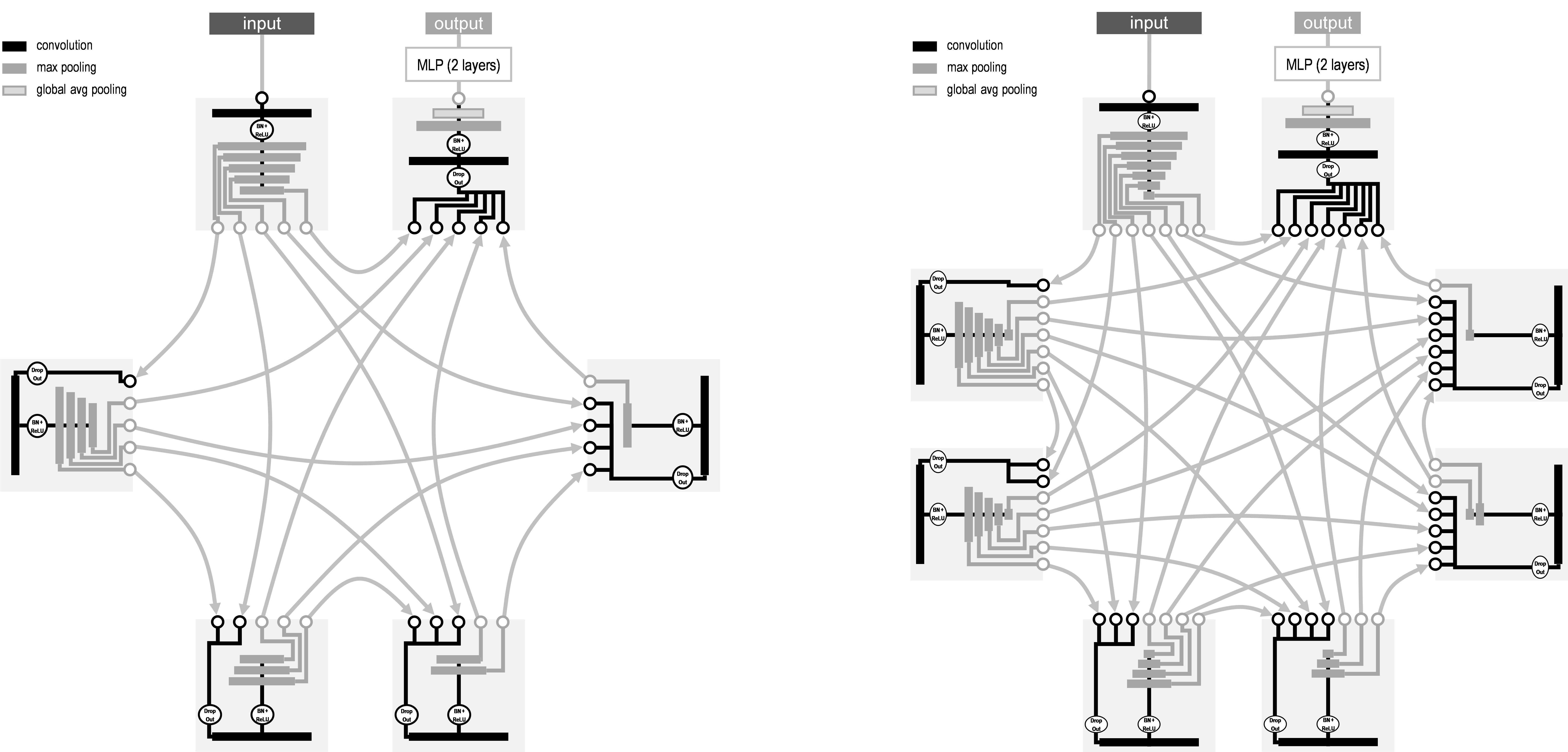}
		\caption{SpiderNet-6 (left) and SpiderNet-8 (right) neural network architectures for fraud detection tasks.}
		\Description{SpiderNet-6 (left) and SpiderNet-8 (right) neural network architectures for fraud detection tasks.}
	\end{figure*}
	\begin{enumerate}
		\item {\itshape Random Forest} is a baseline model that has been chosen as a strong algorithm and industry standard for modeling on tabular data. We used 5-fold cross-validation and the Optuna library to tune the Random Forest hyperparameters \cite{Akiba02}.
		\item {\itshape 1D-CNN} is a one-dimensional convolutional network with classical alternation of convolutional and pooling layers and two fully connected layers and a SoftMax layer. We trained CNN with 3, 6, and 8 convolutional layers;
		\item {\itshape 1D-DenseNet} is a classic DenseNet architecture \cite{Huang22} with an implementation for one-dimensional vectors. We trained two-block architectures with 3 and 4 convolutions in each block;
		\item {\itshape F-DenseNet} is a DenseNet architecture adapted for fraud prediction with two fully connected convolutional blocks containing convolutional and pooling layers. We trained architectures with 3 and 4 convolutional layers in each block;
		\item {\itshape SpiderNet} is our fully connected residual convolutional network with convolutional-pooling blocks. We trained 6 and 8 block architectures (Figure 8).
	\end{enumerate}
	We used weight decay (L2-regularization) in BatchNorm and fully connected layers, and dropout on fully connected layers as regularizers for all neural networks. In the 1D-CNN, F-DenseNet, and SpiderNet architectures we applied the BatchNorm and ReLU transformations after each convolutional layer. In the F-DenseNet and SpiderNet blocks, as an additional regularization for the skip connections, we used dropout after concatenating the input vectors for each block (Figure 4). We also used the fraud-rate leveling technique in batches to solve the problem of the lack of fraud samples in batches in case of over-class imbalance.
	
	To train and assess the quality of the models, we performed a stratified split of private and public datasets into a train $(80\%)$, validation $(10\%)$, and test $(10\%)$ samples. We tuned the networks to the validation sample using Grid-Search and early stopping.
	
	\subsection{Performance Measures}
	We used AUC ROC and AUC PR metrics to evaluate the quality of models. The AUC ROC metric and the linearly related Gini coefficient are industry standards in banking modeling. However, as shown by Saito and Rehmsmeier \cite{Saito45}, the AUC ROC accepts unreasonably high values and becomes uninformative on over unbalanced samples, in contrast to the AUC PR, which adequately estimates the quality of models on unbalanced samples. Therefore, we tuned the hyperparameters of the models using the AUC PR.
	
	To evaluate the quality of fraud detection models on the private dataset, we have developed the special metric PL (Prevented Loss), which shows how much loss from internal fraud the model prevents. To calculate the PL, we estimate prevented loss for each partner:
	
	\begin{equation}
		PL^{\left(i\right)} = P_{\left(T_l - T_a\right)} \cdot \frac{DR - DR_0}{1 - DR_0}
	\end{equation}
	
	\begin{list}{}
		\item where $T_l$ is the whole considered period of loss;
		\item $T_a$ is the period on which the model works;
		\item $\left(T_l - T_a\right)$ is the period after the model is triggered (for our sample, it is 90 days – the empirical period for which the bank's Security detects fraud without using the model);
		\item $DR$ is the Default-Rate for loans issued by the partner for the period $\left(T_l - T_a\right)$;
		\item $DR_0$ is a "zero target" for Default-Rate in which the loan portfolio has zero profit;
		\item $P_{\left(T_l - T_a\right)}$ is the partner's loan portfolio for the period $\left(T_l - T_a\right).$
		
	\end{list}
	
	Total PL is calculated based on the company's total investigative resources, i.e. on the assumption that security costs do not increase. Total Prevented Loss shows the net profit from the model due to the earlier detection of fraud than it had been before the model was applied:
	\begin{equation}
		PL = \sum_{i=1}^{k} PL^{\left(i\right)} \cdot b_i
	\end{equation}
	
	\begin{list}{}
		\item where $PL^{\left(i\right)}$ is prevented loss for the ith fraud partner;
		\item $k$ is the number of the first $k$ partners with the highest model probability of fraud (for our bank $k = 40$);
		\item $b_i$ is a binary variable showing the event for the ith partner: 1 – fraud, 0 – no fraud.
	\end{list}
	For the public dataset, the economic metric was not developed due to the lack of the necessary financial data in the dataset.

	\begin{table*}
		\caption{Quality of models for internal (private dataset) and transactional (public dataset) fraud detection: the best results are highlighted in bold; $95\%$ confidence intervals are shown in parentheses.}
		\label{tab:result}
		\begin{tabular}{ll|ll|ll} 
			
			\textbf{\#} & \textbf{Model} & \multicolumn{2}{c|}{\textbf{Private data (test sample)}} &
			\multicolumn{2}{c}{\textbf{Public data (test sample)}} \\
			\cline{3-6}
			&& \multicolumn{1}{c}{\textbf{AUC PR}} & \multicolumn{1}{c|}{\textbf{AUC ROC}}
			& \multicolumn{1}{c}{\textbf{AUC PR}}
			& \multicolumn{1}{c}{\textbf{AUC ROC}} \cr
			\midrule
			
			1& Random Forest &  0.0650 $(\pm0.001116)$ & 0.9371 $(\pm0.009253)$ & 0.4881 $(\pm0.003114)$ & 0.9709 $(\pm0.003572)$\\
			\hline
			2& CNN-3 & 0.0527 $(\pm0.001012)$ & 0.9339 $(\pm0.012978)$  & 0.4462 $(\pm0.003096)$ &	0.9670 $(\pm0.004674)$\\
			3 & CNN-6 & 0.0644 $(\pm0.001111)$	& 0.9385 $(\pm0.011432)$  & 0.4908 $(\pm0.003114)$	& 0.9711 $(\pm0.004780)$ \\
			4 & CNN-8 & 0.0708 $(\pm0.001161)$ & 0.9288 $(\pm0.009605)$		& 0.5099 $(\pm0.003114)$	& 0.9718 $(\pm0.004511)$ \\
			\hline
			5 & DenseNet-6 [3; 3] & 0.0646 $(\pm0.001113)$ & 0.9315 $(\pm0.009091)$  & 0.4757 $(\pm0.003111)$	& 0.9669 $(\pm0.004935)$ \\
			6& DenseNet-8 [4; 4] & 0.0691 $(\pm0.001148)$ & 0.9310 $(\pm0.010545)$ 	& 0.4854 $(\pm0.003113)$	& 0.9686 $(\pm0.004661)$ \\
			\hline
			7 & F-DenseNet-6 [3; 3] & 0.0732 $(\pm0.001179)$ & 0.9263 $(\pm0.014509)$  & 0.5092 $(\pm0.003114)$	& 0.9708 $(\pm0.005082)$ \\
			8 & F-DenseNet-8 [4; 4] & 0.0575 $(\pm0.001054)$ & 0.9186 $(\pm0.015820)$  & 0.4968 $(\pm0.003114)$	& 0.9704 $(\pm0.004780)$ \\
			\hline
			9 & SpiderNet-6 & \textbf{0.0948} $(\pm0.001326)$ & \textbf{0.9484} $(\pm0.008004)$  & \textbf{0.5375} $(\pm0.003106)$	& \textbf{0.9721} $(\pm0.004763)$ \\
			10 & SpiderNet-8 & 0.0680 $(\pm0.001139)$ & 0.9277 $(\pm0.009588)$  & 0.5160 $(\pm0.003113)$	& 0.9684 $(\pm0.004744)$
		\end{tabular}
	\end{table*}

	\section{Experiment}
	\subsection{Model Optimization and Tuning}
	We tuned models by AUC PR and selected the best hyperparameters for the models.
	\begin{description}[leftmargin=0pt, parsep = 0.7ex, topsep = 1ex]
		\item[\underline{\normalfont Random Forest:}] \hspace{0pt} \\
		{\itshape Private data}: class\_weight=0.0167, max\_depth=7, n\_estimators=129;
		
		{\itshape Public data}: class\_weight=0.0744, max\_depth=7, n\_estimators=90.
		\item[\underline{\normalfont CNN:}] \hspace{0pt} \\
		{\itshape Private data}: layers=8, l2\_batch=0.0001, kernel\_size=3,  dropout=0.25, weight\_decay=0.0002, filters=10, learning\_rate=0.003, hidden=100;
		
		{\itshape Public data}: layers=8, l2\_batch=0.0001, kernel\_size=5,  dropout=0.25, weight\_decay=0.0002, filters=10, learning\_rate=0.003, hidden=30.
		\item[\underline{\normalfont DenseNet:}] \hspace{0pt} \\
		{\itshape Private data}: block\_sizes = [4, 4], initial\_filters=5, initial\_stride=1, k=5, conv\_kernel\_width=3, bottleneck\_size=2, theta=0.5, transition\_pool\_stride=1, initial\_conv\_width=5, initial\_pool\_width=2, initial\_pool\_stride=2;\\
		{\itshape Public data}: block\_sizes = [4, 4], initial\_filters=5, initial\_stride=1, k=10, conv\_kernel\_width=5, bottleneck\_size=2, theta=0.5, transition\_pool\_stride=1, initial\_conv\_width=5, initial\_pool\_width=2, initial\_pool\_stride=2.
		\item[\underline{\normalfont F-DenseNet:}] \hspace{0pt} \\
		{\itshape Private data}: blocks = 2, convolutions=3+3, weight\_decay=0.0002, l2\_batch=0.0002, kernel\_size=7, filters=15, hidden=60, dropout=0.25, learning\_rate=0.003;\\
		{\itshape Public data}: blocks = 2, convolutions=3+3, weight\_decay=0.0002, l2\_batch=0.0002, kernel\_size=5, filters=15, hidden=100, dropout=0.25, learning\_rate=0.003.
		\item[\underline{\normalfont SpiderNet:}] \hspace{0pt} \\
		{\itshape Private data}: blocks=6, l2\_batch=0.0001, kernel\_size=3, filters=10, hidden = 100, weight\_decay=0, dropout=0.25, learn\_rate=0.005, dropout\_block\_k=$0.001*k^4$ (where k=$\{4, 5\}$ is a block number);\\
		{\itshape Public data}: blocks=6, l2\_batch=0.0002, kernel\_size=7, filters=15, hidden=30, weight\_decay=0.0002, dropout=0.25, learn\_rate=0.005, dropout\_block\_k=$0.001*k^4$ (where k=$\{4, 5\}$ is a block number).
	\end{description}
	For all neural networks, we used the Adam optimizer.
	
	\subsection{Results}
	The results of the trained models showed that the SpiderNet-6 architecture had the best quality on both datasets for both AUC PR and AUC ROC (Table 2). Confidence intervals were calculated using the asymptotic method for AUC PR \cite{Boyd69} AUC ROC \cite{Cortes70}. We see clear dynamics that increase in the number of skip-connections results in increase in quality for fraud detection models: {CNN-8: 0 skip connections} $\rightarrow$ {F-DenseNet-6: 2 skip connections} $\rightarrow$ {SpiderNet-6: 10 skip connections}. This confirms our hypothesis that strong features can pass well from the different layers of neural network immediately to its output due to skip connections. On the other hand, we see that the best model of the classic DenseNet-8 performs worse than the best CNN-8, which has no skip connections. This indirectly confirms the hypothesis that, for tabular data, bottlenecks between blocks don't allow strong features to go directly to the output layers of the network. The F-DenseNet architecture, which is adapted for fraud detection tasks, does not contain a bottleneck between blocks, so the quality of the best F-DenseNet-6 surpasses the quality of CNN-8 and DenseNet-8. These results demonstrate the importance of developing a neural network architecture for the specifics of the task.
	
	\begin{table}
		\caption{Quality of models for internal fraud detection.}
		\label{tab:result2}
		\begin{tabular}{ll|cc} 
			
			\textbf{\#} & \textbf{Model} & \multicolumn{2}{c}{\textbf{Private data (test sample)}}\\
			\cline{3-4}
			&& \multicolumn{1}{c}{\textbf{Fraud, \#}} & \multicolumn{1}{c}{\textbf{PL}} \cr
			\midrule
			\mathstrut & Random classifier & 48 & \textdollar325 604 \\
			\hline
			1& Random Forest & 208 & \textdollar2 079 527 \\
			\hline
			2& CNN-3 & \textbf{312} & \textdollar2 235 707 \\
			3 & CNN-6 & 280 & \textbf{\textdollar2 753 821} \\
			4 & CNN-8 & 280 & \textdollar2 337 297 \\
			\hline
			5 & DenseNet-6 [3; 3] & 240 & \textdollar2 324 181 \\
			6& DenseNet-8 [4; 4] & 288 & \textdollar2 433 914 \\
			\hline
			7 & F-DenseNet-6 [3; 3] & 240 & \textdollar2 297 848 \\
			8 & F-DenseNet-8 [4; 4] & 272 & \textdollar2 402 470 \\
			\hline
			9 & SpiderNet-6 & 304 & \textdollar2 570 014 \\
			10 & SpiderNet-8 & 264 & \textdollar2 379 977 \\
			\hline
			\mathstrut & Perfect classifier & 888 & \textdollar4 659 439 
		\end{tabular}
	\end{table}
	
	\begin{table}
		\caption{Sign tests for two pairs of results (the PL and Fraud metrics are normalized according to the perfect classifier).}
		\label{tab:result3}
		\begin{tabular}[t]{p{0.21\linewidth}|p{0.15\linewidth}p{0.15\linewidth}|p{0.15\linewidth}p{0.15\linewidth}}
			
			\mathstrut & \textbf{CNN-3} & \textbf{Spider \linebreak Net-6} & \textbf{CNN-6} &	\textbf{Spider \linebreak Net-6}\\
			
			\hline
			\textbf{Private data:} & \mathstrut & \mathstrut & \mathstrut & \mathstrut \\
			AUC PR& 0.0527 & \textbf{0.0948} & 0.0644 &  \textbf{0.0948}\\
			AUC ROC& 0.9339 & \textbf{0.9484} & 0.9385 & \textbf{0.9484} \\
			PL (recall)& 0.4798 & \textbf{0.5516} & \textbf{0.5910} & 0.5516\\
			Fraud (recall)& \textbf{0.3514} & 0.3423 & 0.3153 & \textbf{0.3423}\\
			\textbf{Public data:} & \mathstrut & \mathstrut & \mathstrut & \mathstrut \\
			AUC PR& 0.4462 & \textbf{0.5375} & 0.4908 & \textbf{0.5375}\\
			AUC ROC& 0.9670 & \textbf{0.9721} & 0.9711 & \textbf{0.9721}\\
			\hline
			p-value& \multicolumn{2}{c|}{0.015625} &  \multicolumn{2}{c}{0.015625}\\
		\end{tabular}
	\end{table}
	We also confirm Saito’s and Rehmsmeier’s results \cite{Saito45} that the AUC ROC metric is not informative for problems with unbalanced data and shows unreasonably high values. For our datasets, the AUC PR metric is preferred. Although the results for the private dataset according to the AUC PR metric are low, we consider them satisfactory, since they don't take into account the full effect of the models, but demonstrate an uplift when switching from the rule-based approach to model one. In addition, the evaluation of these models according to the PL metric showed a significant positive economic effect with an unchanged budget for investigations (Table 3). Since the PL metric is not normalized, Table 3 also shows the efficiency for the random and perfect classifiers on the private dataset. The results demonstrate that the SpiderNet has an almost 8-fold increase in financial efficiency compared to a random classifier and has less than a 2-fold decrease to the perfect model (as if the bank had identified all fraudulent POS partners before committing crimes – so-called "Minority Report").
	
	We also see that for the private dataset, SpiderNet-6 ranked second according to PL metric, losing to the simpler CNN-6. This may be because the hyperparameters in the Grid-Search were selected following the integral metric AUC PR, while the PL metric is a threshold and covers a small number of POS-partners who have been checked by the bank's security (40 partners per week out of 6786 on average).
	
	On the other hand, as we can see from the results of Table 3, SpiderNet-6 outperforms CNN-6 by the number of identified fraudulent POS partners (out of all limited number of those verified by the security) but loses to CNN3 by this indicator. The number of fraudulent partners detected is also an important metric of the model quality. Hand et al. \cite{Hand71} suggest that in mass fraud, it is better to tune the model to maximize the number of detected fraud cases than to maximize prevented losses. This approach can have a greater influence on fraudsters, since, as we know from Becker's concept, the number of identified fraudulent cases directly proportionally affects the probability of being caught (eq. 1).
	The results obtained show that on various datasets and various quality metrics, SpiderNet-6 loses to CNN-3 and CNN-6 only in one metric. At the same time, in 5 other cases, SpiderNet-6 wins over CNN-3 and CNN-6. To assess the statistical significance of the obtained SpiderNet-6 advantage, we tested the null hypothesis and alternative one-sided hypothesis using a sign test \cite{Conover72}:
	\begin{enumerate}
		\item For SpiderNet-6 and CNN-3
		
		$H_0^{(1)}:$ models are of the same quality
		\begin{displaymath}
			P(Quality_{ SpiderNet-6} > Quality_{ CNN-3}) = 0.5
		\end{displaymath}
		$H_1^{(1)}:$ SpiderNet-6 has higher quality
		\begin{equation*}
			P(Quality_{ SpiderNet-6} > Quality_{ CNN-3}) > 0.5
		\end{equation*}
		\item For SpiderNet-6 and CNN-6
		
		$H_0^{(2)}:$ models are of the same quality
		\begin{equation*}
			P(Quality_{ SpiderNet-6} > Quality_{ CNN-6}) = 0.5
		\end{equation*}
		
		$H_1^{(2)}:$ SpiderNet-6 has higher quality
		\begin{equation*}
			P(Quality_{ SpiderNet-6} > Quality_{ CNN-6}) > 0.5
		\end{equation*}
	\end{enumerate}
	
	\begin{figure}[ht]
		\centering
		
		\centering\includegraphics[width=\linewidth]{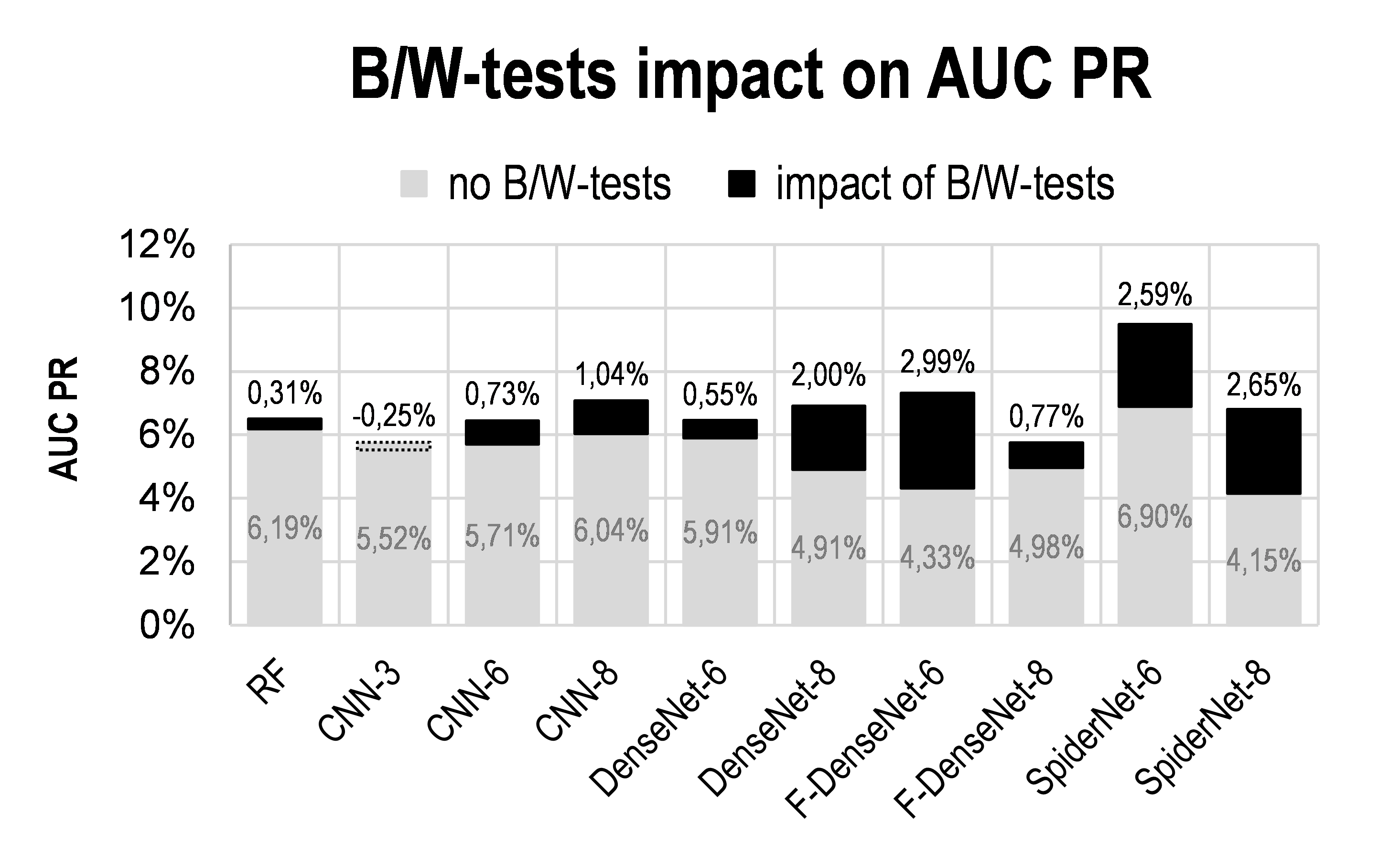}
		
		\caption{Influence of B-tests and W-tests on model quality (negative impact means a decrease in the model quality when adding B/W-tests)}
		\Description{Influence of B-tests and W-tests on model quality (negative impact means a decrease in the model quality when adding B/W-tests)}
	\end{figure}
	
	\begin{figure}[ht]
		\centering
		\begin{subfigure}[b]{0.5\linewidth}
			\centering\includegraphics[width=\linewidth]{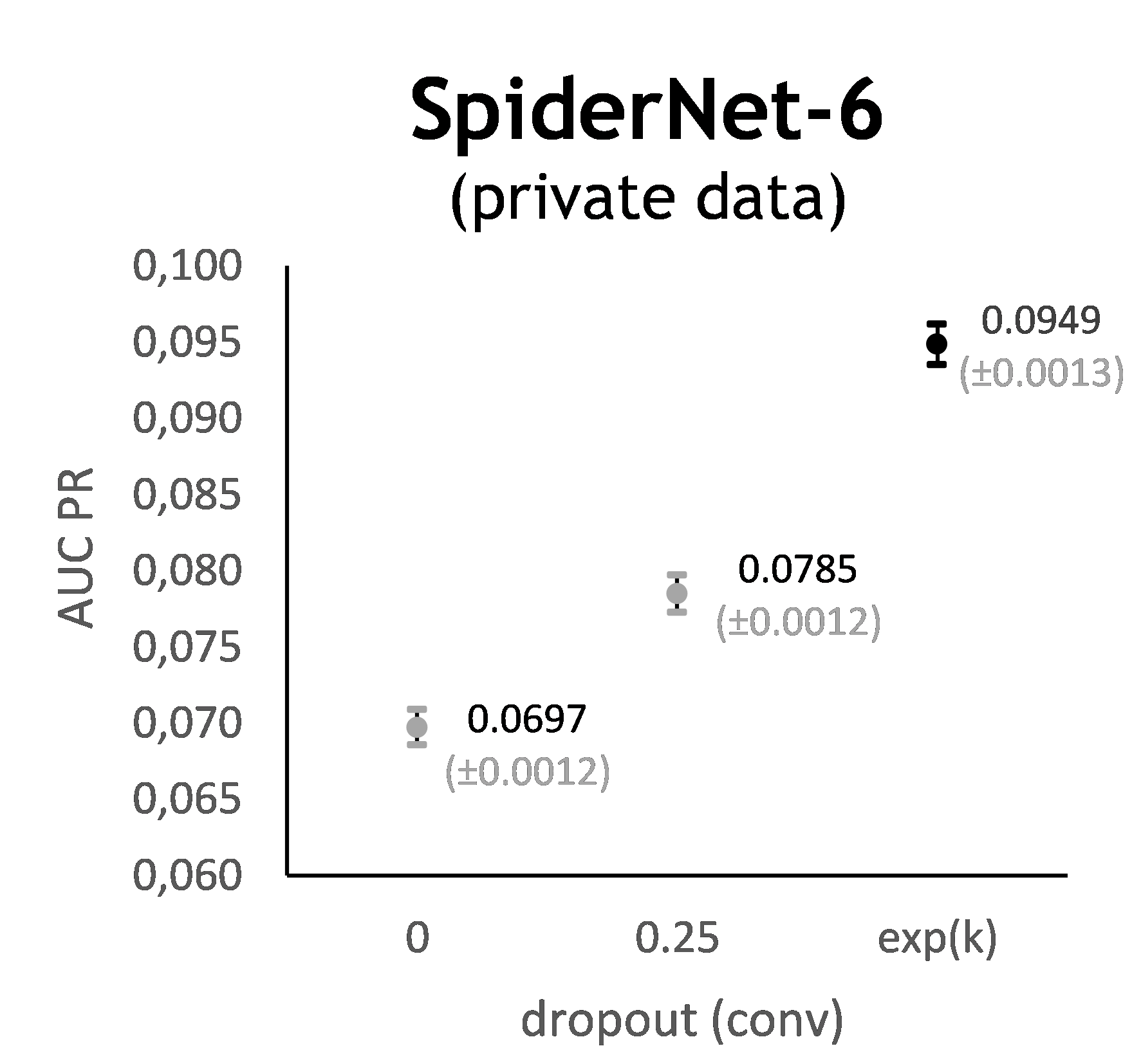}
		\end{subfigure}%
		\begin{subfigure}[b]{0.5\linewidth}
			\centering\includegraphics[width=\linewidth]{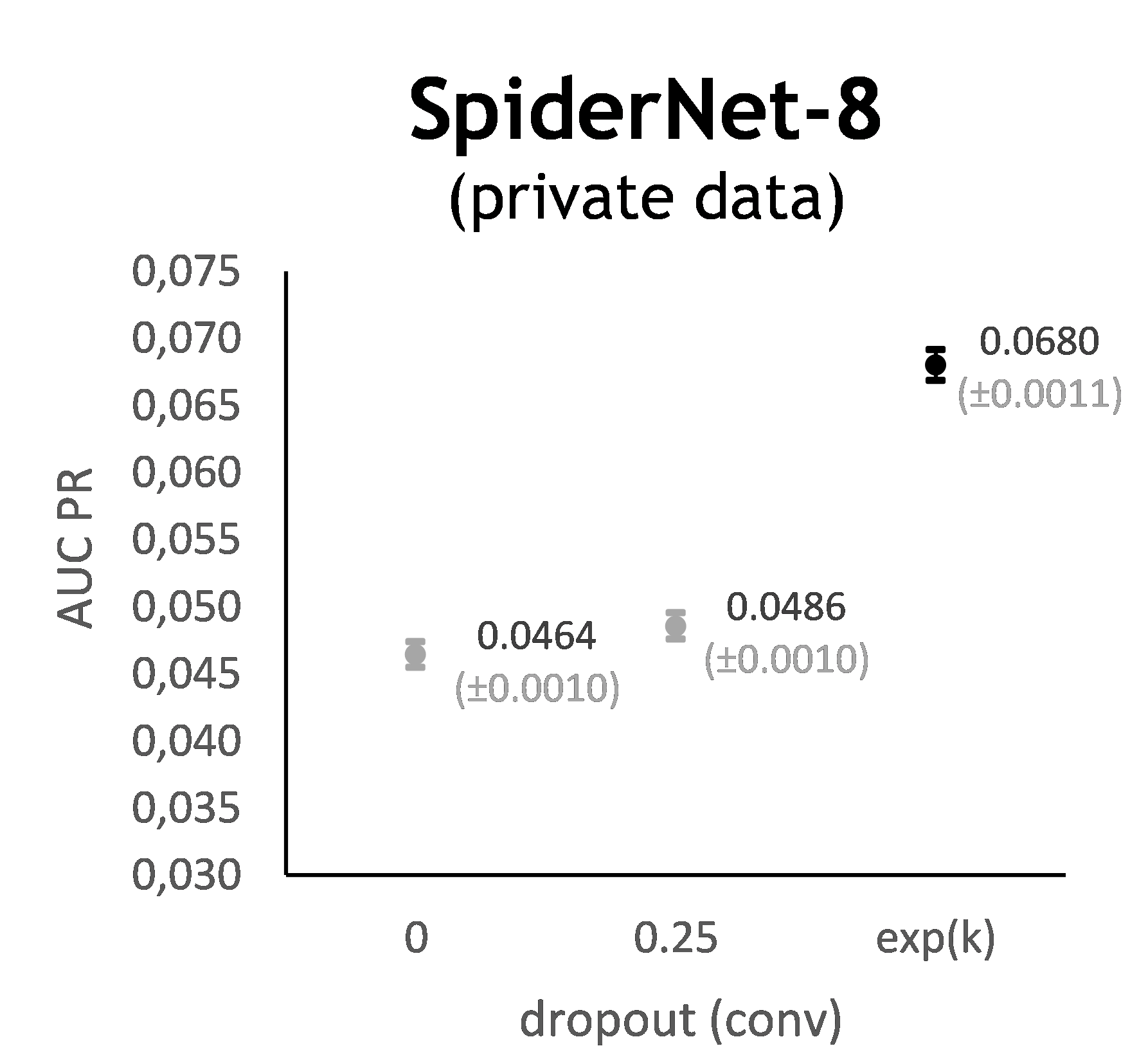}
		\end{subfigure}\vspace{0pt}
		
		\begin{subfigure}[b]{0.5\linewidth}
			\centering\includegraphics[width=\linewidth]{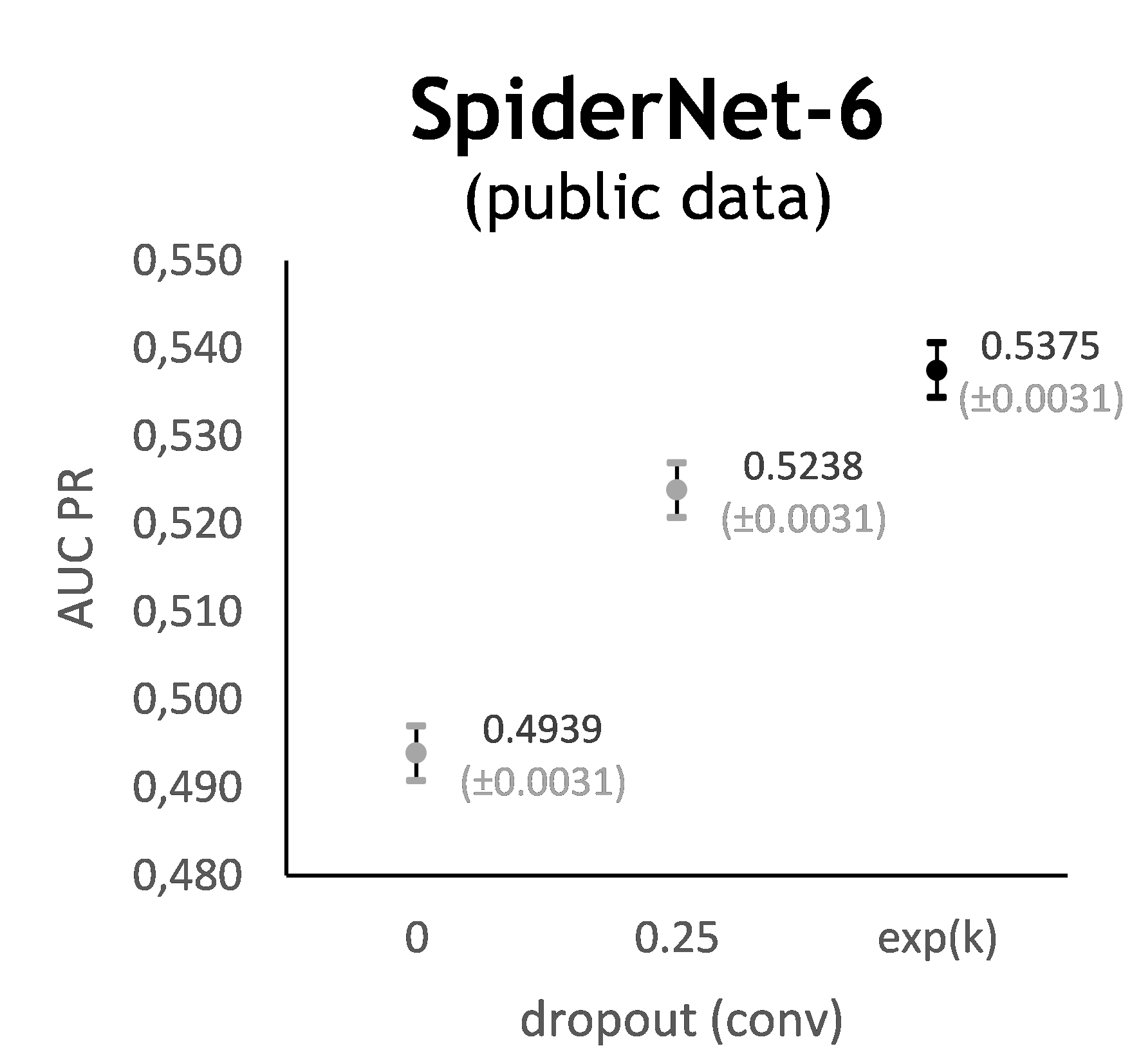}
		\end{subfigure}%
		\begin{subfigure}[b]{0.5\linewidth}
			\centering\includegraphics[width=\linewidth]{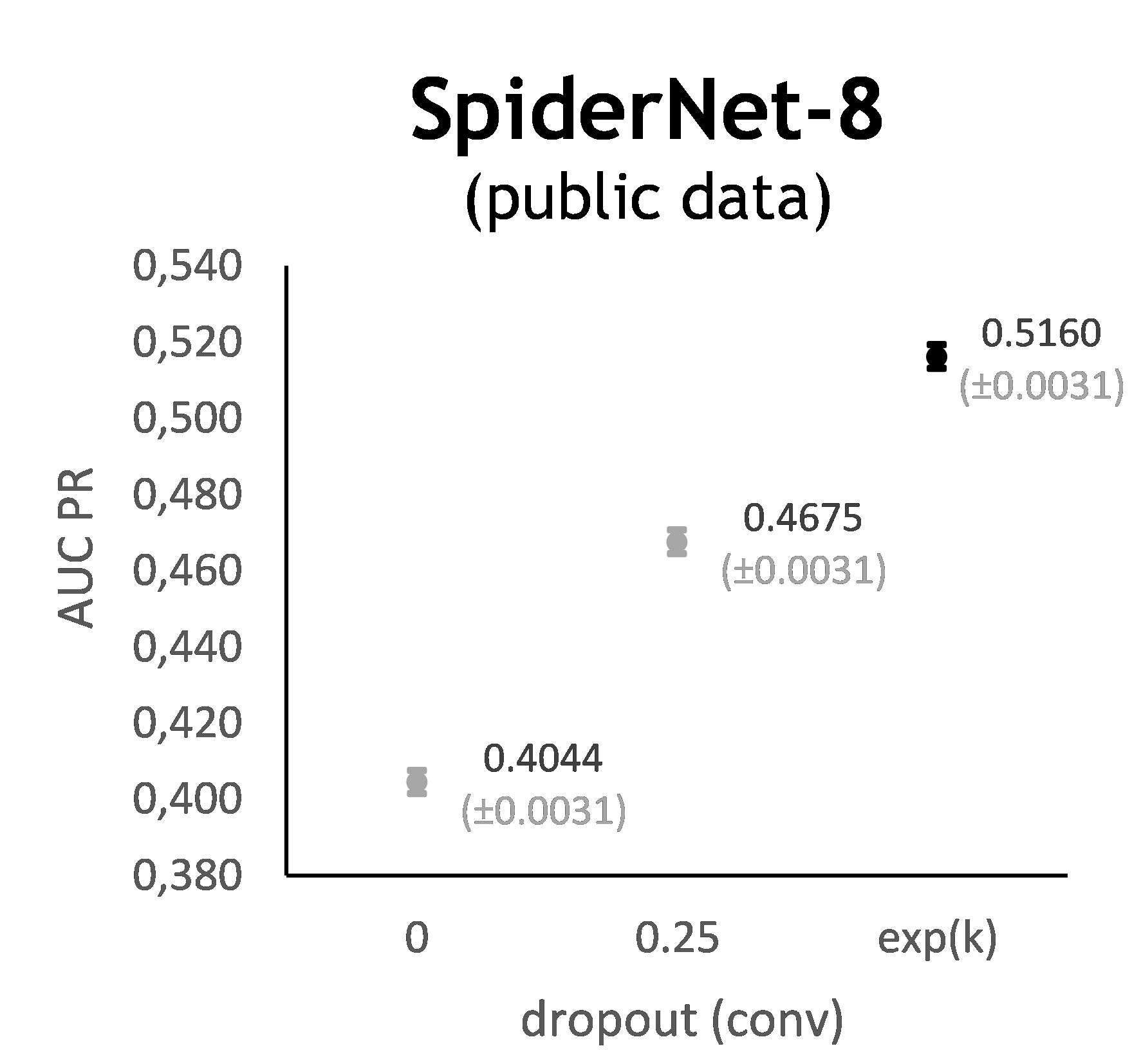}
		\end{subfigure}
		\caption{SpiderNet quality (AUC PR), depending on the dropout techniques in convolutional layers: 1) the value of zero corresponds to SpiderNet quality without a dropout; 2) the value of 0.25 corresponds to constant dropout=0.25 used in all Spider-blocks (vanilla technique); 3) the value of exp(k) corresponds to an exponential increase in the dropout value as the Spider-block number increases (our technique).}
		\Description{SpiderNet quality.}
	\end{figure}
	The obtained results of p-values are presented in Table 4 and we see that in both pairwise comparisons the null hypothesis is rejected in favor of the alternative, i.e. with a 95\% significance level SpiderNet-6 more often shows better quality than CNN-3 or CNN-6.
	
	We also tested the effectiveness of B-tests and W-tests by training the models on two private samples:
	\begin{enumerate}
		\item Full sample containing 24 B-tests, 15 W-tests, and 124 expert rules;
		\item Truncated sample containing 124 expert rules.
	\end{enumerate}
	
	Our results showed the high efficiency of B-tests and W-tests by AUC PR (Figure 9). Adding B-tests and W-tests gave us the AUC PR increase of $19.4\%$. At the same time, we see that B/W-tests have a greater impact on models with skip connections. This may be because B/W-tests are strong rules and for a better impact they must be sent immediately to the network output.
	
	We also see that SpiderNet-6 trained on expert rules without B-tests shows the best quality by AUC PR metric, which once again proves the stability of the results obtained.
	
	In addition, the results of experiments demonstrated that our exponential dropout technique in convolutional layers provides significant gains in quality over a constant or zero dropout (Figure 10). This trick allows adding more regularization in the last blocks, which get more information from the previous blocks.
	
	All program codes and detailed results are available at: \url{https://github.com/aasmirnova24/SpiderNet}
	
	\section{Conclusions}
	In this paper, we investigated deep learning methods for fraud prediction and proposed a novel architecture of neural networks for fraud detection problems. Taking inspiration from the skip connection concepts of Resnet \cite{He16}, FractalNet \cite{Larsson30}, Adanet \cite{Cortes10}, and DenseNet \cite{Huang22} convolutional networks for computer vision, we developed SpiderNet architecture for fraud detection, guided by anti-fraud expert knowledge. Using convolutional layers, our SpiderNet creates hierarchical combinations of anti-fraud rules, and a fully connected structure of skip connections between blocks allows strong rules to be forwarded immediately to the network output. Also, our network can select strong rules early on through the use of a multi-layered structure of pooling layers in Spider-blocks. Moreover, the down-dropout technique we use between Spider-blocks is an additional regularizer against the excessive complexity of the network.
	
	In our opinion, SpiderNet should work well for heterogeneous input data, when there are clear leaders among the rules supplied to the network input and they must be forwarded to the output without additional transformation (for example, scores of other models or strong rules).
	
	It can also be noted that our results confirm the hypothesis of ResNet authors, which they formulated in their paper \cite{He16}: "The residual learning principle is generic, and we expect that it is applicable in other vision and non-vision problems".
	
	In this paper, we also proposed new methods for developing anti-fraud rules – B-tests and W-tests, which significantly affect the quality of the models. We hope that B/W-tests will contribute to the solution of the problem “The Blind Men and the Elephant”, which was formulated by Foster Provost \cite{Bolton06}.
	
	We also hope that the metric of quality Prevented Losses developed by us allows us to solve an important issue for the industry: how to evaluate the economic efficiency of anti-fraud models developed for industrial use.
	
	We understand that our SpiderNet does not solve all the anti-fraud modeling problems identified by Bolton, Hand, Provost, and Breiman \cite{Bolton06}. In particular, we still use expert rules designed for specific fraud types to train models. Our B-tests and W-tests partially solve this problem, but there are other strong methods for fraud feature engineering, such as graph methods \cite{Shuhan47, Wang55, Dou12, Wang56}, entropy changing methods \cite{Fu14}, and variance anomaly detection methods (V-tests, similar to B-tests). We plan to work on these topics in our future research.
	
	An important component of SpiderNet is skip-connection, which helps to forward strong features directly to the output layers of the network, partially solving the problem of locality in convolutions, when the order of features in the input vector is important, and their rearrangement leads to a change in the quality of the model. However, the current implementation of SpiderNet does not completely solve the locality problem. Our future work will focus on this problem.
	
	We also assume that SpiderNet might work well not only for fraud detection but also for other types of modeling tasks that use tabular data. Testing this hypothesis is also a topic for our future research.
	
	\begin{acks}
		We are grateful to Dmitry Efimov for his valuable advice.
	\end{acks}


	\bibliographystyle{ACM-Reference-Format}
	\bibliography{bib_full_auth.bib}

\end{document}